\documentclass[letterpaper, 10 pt, conference]{ieeeconf}

\IEEEoverridecommandlockouts                              

\usepackage{amsmath} 

\usepackage{algorithm,algpseudocode}
\usepackage{amsfonts}
\usepackage{array}
\usepackage{tabularx}

\usepackage{amssymb}

\title{\LARGE \bf 3D Laser-and-tissue Agnostic Data-driven Method for Robotic Laser Surgical Planning}

\author{Guangshen Ma $^{1}$, Ravi Prakash $^{1}$, Brian Mann, Ph.D $^{1}$, Weston Ross, Ph.D $^{2}$ and Patrick Codd, M.D $^{1, 2}$ 
\thanks{$^{1}$ Department of Mechanical Engineering and Materials Science, Duke University. Corresponding author: guangshen.ma@duke.edu. }%
\thanks{$^{2}$ Department of Neurosurgery, Duke University Medical Center.}
}

\usepackage{lipsum}
\usepackage{graphicx}

\usepackage{multirow}

\usepackage{lipsum}
\usepackage{adjustbox}
\usepackage{comment}
\usepackage{xcolor}
\usepackage{float}
\usepackage{caption}
\usepackage{subcaption}
\usepackage{tabularx}
\usepackage[english]{babel}
\usepackage[utf8]{inputenc}
\usepackage{algorithm}
\usepackage{amsmath}
\usepackage{hhline}

\usepackage{nth}

\usepackage{caption}
\usepackage{booktabs}

\usepackage{algorithm}
\usepackage{algpseudocode}
\usepackage{listings}

\usepackage{amssymb}

\begin{document}

\maketitle
\thispagestyle{empty}
\pagestyle{empty}

\begin{abstract}

In robotic laser surgery, shape prediction of an one-shot ablation cavity is an important problem for minimizing errant overcutting of healthy tissue during the course of pathological tissue resection and precise tumor removal. Since it is difficult to physically model the laser-tissue interaction due to the variety of optical tissue properties, complicated process of heat transfer, and uncertainty about the chemical reaction, we propose a 3D cavity prediction model based on an entirely data-driven method without any assumptions of laser settings and tissue properties. Based on the cavity prediction model, we formulate a novel robotic laser planning problem to determine the optimal laser incident configuration, which aims to create a cavity that aligns with the surface target (e.g. tumor, pathological tissue). 

To solve the one-shot ablation cavity prediction problem, we model the 3D geometric relation between the tissue surface and the laser energy profile as a non-linear regression problem that can be represented by a single-layer perceptron (SLP) network. The SLP network is encoded in a novel kinematic model to predict the shape of the post-ablation cavity with an arbitrary laser input. To estimate the SLP network parameters, we formulate a dataset of one-shot laser-phantom cavities reconstructed by the optical coherence tomography (OCT) B-scan images for the data-driven modelling. To verify the method. The learned cavity prediction model is applied to solve a simplified robotic laser planning problem modelled as a surface alignment error minimization problem. The initial results report $(91.1 \pm 3.0)\%$ 3D-cavity-Intersection-over-Union (3D-cavity-IoU) for the 3D cavity prediction and an average of $97.9\%$ success rate for the simulated surface alignment experiments.

\end{abstract}

\section{INTRODUCTION}

Robotic laser systems have been used to provide accurate and rapid control of surgical laser-scalpels in various medical applications such as eye surgery \cite{MonocularLaser}, neurosurgery \cite{ross2018automating, liao2012integrated} and dermatology \cite{wheeland1995clinical}. A laser incident configuration can be described by a 6-degree-of-freedom (dof) model which encodes the 3-dof laser ablation incident center and the 3-dof laser orientation vector \cite{li1995laser}. Different laser configurations can create various tissue ablation cavities with unique shapes. An incorrect laser configuration can cause over-irradiation of healthy tissue that should not be ablated, and thus simulating the shape of the cavity before the actual laser ablation is a necessary step for ensuring a safe, controlled resection of only the target pathological tissue. This brings the need to formulate a cavity prediction model from a given laser incident configuration. 

\begin{figure}[h]
\centering
\includegraphics[scale = 0.45]{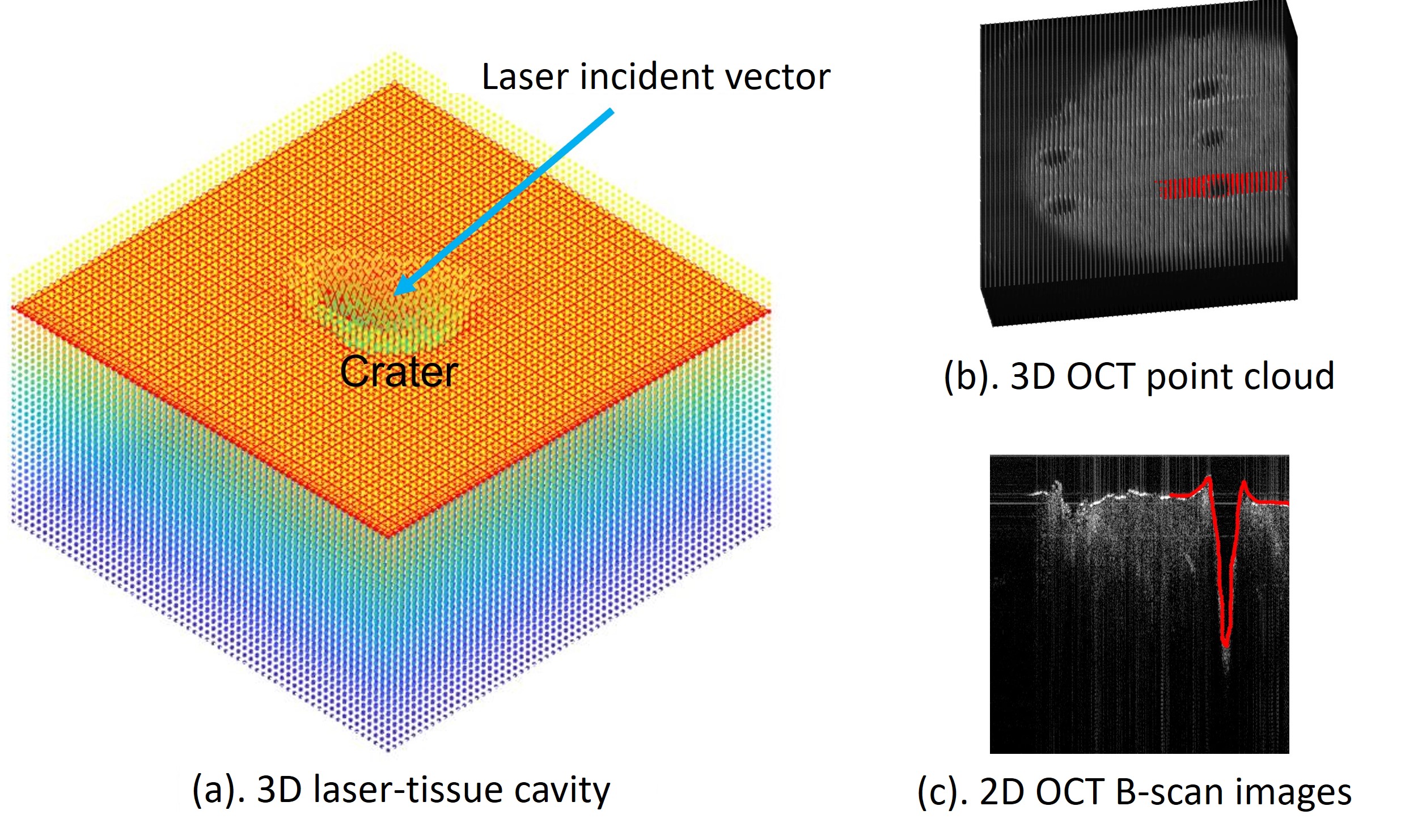}
\caption{One-shot laser cavity prediction on a 3D surface. The laser-tissue cavity is generated from a unique robotic laser configuration (ablation center and incident orientation). (a): an example of a volumetric point cloud with a superficial surface cavity. (b): a 3D volume from the OCT B-scan images. (c): a B-scan image of a labelled 2D-cavity.}
\label{fig_1_oct_3d_cavity}
\end{figure}

Additionally, the cavity prediction model can be applied in solving the robotic laser surgical planning problem. This problem aims to determine an optimal laser incident configuration to create an ablation cavity aligned with the surface target, such as tumor, pathological tissue boundary (Fig.~\ref{fig_1_oct_3d_cavity}a). Before the actual laser ablation, the superficial layer of the tissue can be extracted from the OCT B-scan images and a 3D intraoperative surface can be formulated based on post-processing computer vision approaches \cite{draelos2021contactless}. It is important to predict the shape of the post-ablation cavity to minimize the over-cutting and under-cutting of the pathological tissue. In summary, we address two research problems as:
\begin{enumerate}
    \item Laser-tissue cavity prediction: How to develop a cavity prediction model that can map a laser incident configuration to a predicted surface cavity?
    \item Robotic laser surgical planning: Given a surface target, how to determine an optimal laser incident configuration to create an ablation cavity, such that the offset between the two profiles can be minimized? 
\end{enumerate}

\begin{figure*}[h]
\centering
\includegraphics[scale = 0.53]{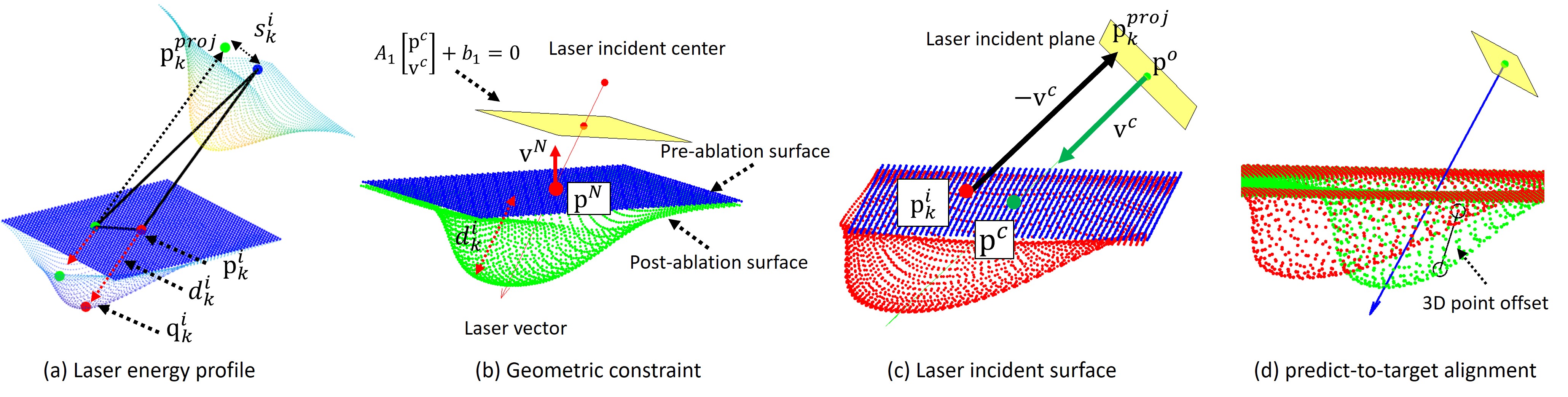}
\caption{Laser-tissue geometric configurations showing: (a) the laser energy profile and the resulted output layout of the surface, (b) the depth-of-cut between the pre- and post- surfaces, and the equality constraint of the incident center denoted as a planar surface, (c) the laser incident plane with the laser origin $\mathbf{p}^o$, the incident vector $\mathbf{v}^c$ and the ablation center $\mathbf{p}^c$, and (d) two predicted cavities from difference laser incident configurations and the two point correspondences (same index).}
\label{fig_laser_tissue_geo_config}
\end{figure*}

\subsection{Data-driven Laser-tissue Cavity Prediction} 

Modelling the laser-tissue interaction as a dynamic system is a difficult problem due to the heterogeneity of tissue material and the complex physical mechanism \cite{fichera2016online, kahrs2010planning}. Therefore, there is a need to use a data-driven method to directly learn the geometric relation between the laser energy profile on the surface and the corresponding deformation. The data-driven method has the benefit of encoding the physical mechanism into a model with parameters learned by the data directly, which has been widely applied in the field of physics-inspired robot learning \cite{zhu2019data}.

The conventional methods for laser tissue photoablation modelling follow the Beer-Lambert Law \cite{niemz2007laser} to predict the ablation depth of cut \cite{ross2018automating}. However, this model requires the prior knowledge of the tissue absorption coefficient ($\mu_a$), the incident laser intensity ($\Phi_0$), and the tissue ablation enthalpy ($\Phi_{th}$) \cite{Vogel2003, ross2018optimized}. In most medical applications, the tissue properties can not be easily determined and the laser setting can be varied depending on the particular use. There is a need to develop a method to automatically fine-tuned with the unique tissue-laser properties. Another method to model the laser-tissue interaction follows the Monte Carlo method \cite{wang1992monte} to simulate the absorption and scattering of the light propagation in tissue. However, this method cannot be directly applied to modelling the surface profile in 3D without accurate knowledge of the tissue optical properties and is computationally prohibitive for real-time implementation.

The recent development with data-driven methods opens new research directions for solving the laser-tissue cavity modelling problem. The shape of the resulting ablation crater is generally assumed to be similar to the Gaussian beam profile since the depth-of-cut is related to the strength of energy delivered to the target \cite{fichera2016online, ross2018optimized, burgner2010ex}. Therefore, the 3D geometric laser-tissue interaction can also be described by the symmetric Gaussian function to mimic the Gaussian-beam profile property. The parameters of these functions can be learned through the 3D cavity data collected by high-resolution scanners such as confocal microscopy \cite{kahrs2010planning, kahrs2008visual} and computed tomography (CT) \cite{ma2020characterization}. However, there is a limitation on using a symmetric Gaussian function to model the laser-tissue cavity since the actual ablation craters do not usually follow the perfect assumption of Gaussian profiles. In addition, these studies have not discussed the problem of laser surgical planning and the application of controlling the ablated profiles for robotic laser surgery. 

\subsection{Robotic Laser Surgical Planning} 

An incorrect laser incident configuration can create an ablation cavity that is not aligned with the surface target, e.g. volumetric tumor resection. The development of an optimal laser planner not only can help with the precise tumor resection but also minimize the probability of over-cutting the healthy tissue \cite{ma2022robotic}. The robotic laser planning can be modelled as a surface alignment problem, i.e. the minimization of the offset between the predicted cavity and the surface target. The surface targets can mimic the shapes of tumor boundaries and pathological tissue. The optimal laser planning can show applications of precise tissue removal and improving the automated robotic laser surgery. 

\noindent
\textbf{Main Contributions:} The two research contributions are:
\begin{enumerate}
    \item The development of a laser-and-tissue agnostic data-driven method to predict the shape of a 3D cavity from a given robotic laser configuration. 
    \item Applying the proposed cavity prediction model to solve a simplified robotic laser planning problem. 
\end{enumerate}

\section{METHODS}

This section discusses the data-driven method of the laser-tissue cavity prediction modelled as a kinematic system, and the formulation of the optimal robotic laser planning problem using the cavity prediction model.

\subsection{Data-driven Laser-tissue Kinematics} 

Kinematics normally refers to the relation between the robot configuration (system variables) and the output layout, e.g. position and orientation of end-effector of a robot arm \cite{sciavicco2001modelling}. In this study, we re-define the laser-tissue interaction as a kinematic model to map a given laser incident configuration to an ``output profile" of the one-shot laser ablation cavity. Each pre-ablation surface point can be assigned with a depth-of-cut towards the laser incident vector and center, and the added deformation can be concatenated to formulate the post-ablation cavity, as shown in Fig.~\ref{fig_laser_tissue_geo_config}a. Therefore, the laser-tissue interaction can be considered as a ``robot system" and we define the forward and inverse kinematics as:
\begin{itemize}
    \item Forward kinematics (FK): Given a laser incident angle $\mathbf{v}^c \in \mathbb{R}^3$ and a laser ablation center $\mathbf{p}^{c} \in \mathbb{R}^3$, a query point at the pre-ablation surface can be uniquely mapped to a new position $\mathbf{q}^i \in \mathbb{R}^3$ ($\mathbf{q}^i$ corresponds with $\mathbf{p}^i$) via the ablation process. The summation of $\mathbf{q}^i$ formulates the ablation cavity.
    \item Inverse Kinematics (IK): Given a target position $\mathbf{p}^* \in \mathbb{R}^3$, calculate the optimal laser orientation $\mathbf{v}^c$ and ablation center $\mathbf{p}^c$ such that the new position $\mathbf{q}^i$ can have minimal distance to $\mathbf{p}^*$. 
\end{itemize}

We define a tissue surface before laser ablation as the ``pre-ablation surface" $\{\mathbf{p}_k^i\}_{k=1}^M, \mathbf{p}_k^i \in \mathbb{R}^3$ and the one after cutting as ``post-ablation surface" $\{\mathbf{q}_k^i\}_{k=1}^M, \mathbf{q}_k^i \in \mathbb{R}^3$ (the index $k$ refers to $k$-th (arbitrary) point of a surface and $i$ refers to a symbol of a surface point), as shown in Fig.~\ref{fig_laser_tissue_geo_config}b. It is noted that $\{\mathbf{p}_k^i\}_{k=1}^M$ and $\{\mathbf{q}_k^i\}_{k=1}^M$ have the same number of points. Each pair of $\mathbf{p}_k^i$ and $\mathbf{q}_k^i$ formulates a unique correspondence. 

\noindent
\textbf{Forward kinematics:} For a pre-ablation surface point $\mathbf{p}^i_k$:
\begin{equation}
    \begin{aligned}
        \mathbf{q}_k^i (\mathbf{v}^c, \mathbf{p}^c, \mathbf{p}^i_{k}) &  = \mathbf{p}_k^i + d_k^i (\cdot) * \mathbf{v}^c 
    \end{aligned}
    \label{fk_1}
\end{equation}
Where $d_k^i (\cdot) \in \mathbb{R}$ is the depth of cut of $\mathbf{p}_k^i$ from the pre-ablation surface, along the incident orientation $\mathbf{v}^c$. It shows how much tissue can be removed from the $\mathbf{v}^c$. The $\mathbf{q}^i_k$ is the resulting position coordinate updated from $\mathbf{p}^i_k$.

To determine the depth of cut of $\mathbf{p}_k^i$, we use the property of the laser Gaussian beam profile where the point closer to the ablation center shows greater depth-of-cut while the farther point shows less value \cite{Vogel2003, ross2018optimized}. We define a ``distance-to-laser-center" $s_k^i \in \mathbb{R}$ as a metric to measure the offset towards the laser origin and the coordinate projected from a surface point to the laser incident plane, as depicted in Fig.~\ref{fig_laser_tissue_geo_config}a and Fig.~\ref{fig_laser_tissue_geo_config}c. The $s_k^i$ can be calculated by projecting a surface point to the laser incident plane with a unique laser incident vector and origin (Fig.~\ref{fig_laser_tissue_geo_config}c). Therefore, the $d_k^i$ and $s_k^i$ formulate a unique end-to-end correspondence and we can use a function to model this relation.

\begin{figure}[h]
\centering
\includegraphics[scale = 0.26]{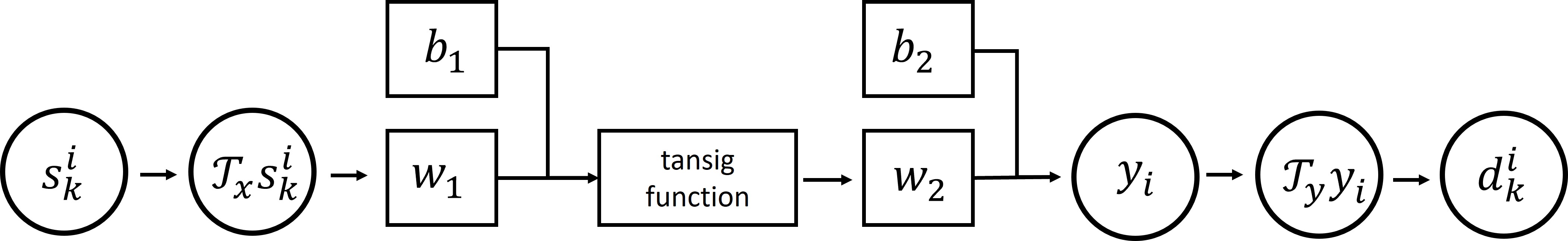}
\caption{Single-layer perceptron (SLP) model configuration. There are four parameters with SLP for data fitting: $w_1$ and $w_2$ are the weights. $b_1$ and $b_2$ are the bias. The $\mathcal{T}_x$ and $\mathcal{T}_y$ are the pre- and post- transformations of the input-output data.}
\label{fig_slp_config}
\end{figure}

\noindent
\textbf{Single layer perceptron (SLP) for data-driven modelling:} A SLP was used to model the one-to-one mapping between $d_k^i$ and $s_k^i$. Our prior works have used the symmetric gaussian-based model to describe the relation between the laser energy and surface geometry \cite{burgner2010ex}. However, the symmetric Gaussian-based model has only two parameters and cannot be generalized to a more complicated model for data fitting. Another option is to use other non-linear regression functions, such as polynomial functions and mixture of gaussian functions. However, these approaches require some specified definitions, such as the order of the polynomial function. Thus, we used a single-layer perceptron (SLP) model with four parameters (the single hidden layer and the output layer contain two parameters for the weights and the bias) as a regression function for the model fitting. SLP is one of the multilayer perceptron (MLP) networks with only one hidden layer. It is noted that SLP model has the benefit of minimizing the possibility of model over-fitting which is common in MLP models when used to fit simple data. With different laser-tissue models, the SLP model parameters can be automatically adjusted for the given datasets. Specifically, we have the SLP model:
\begin{equation}
    \begin{aligned}
        d_k^i (\cdot) & = f_{SLP}( s_k^i ) 
    \end{aligned}
    \label{fk_2}
\end{equation}
Where $f_{SLP}: \mathbb{R} \rightarrow \mathbb{R}$ denotes the SLP as depicted in Fig.~\ref{fig_slp_config}. 

\noindent
\textbf{SLP model encoded in kinematics:} Estimating the parameters of $f_{SLP}$ can be considered as a 1D regression problem between the independent variable $s_k^i$ and the dependent variable $d_k^i$. To compute $s_k^i$, we first define a reference plane to describe the Gaussian beam profile as a point-vector configuration $(\mathbf{p}^o, \mathbf{v}^c)$, which is referred as ``laser incident plane" (Fig.~\ref{fig_laser_tissue_geo_config}a and Fig.~\ref{fig_laser_tissue_geo_config}c). Firstly, we need to project the laser ablation center $\mathbf{p}^c$ to the laser origin $\mathbf{p}^o$ at the laser incident plane:
\begin{equation}
    \begin{aligned}
         \mathbf{p}^o & = \mathbf{p}^c - L_{ref} * \mathbf{v}^c
    \end{aligned}
    \label{pts_from_center}
\end{equation}
where $L_{ref} = 1$ is an arbitrary constant value that is fixed. To get $s_k^i$, we need to know the distance between the projected coordinate at the laser incident plane to the center of the laser origin (Fig.~\ref{fig_laser_tissue_geo_config}c). Given a query point $\mathbf{q}_k^i$, the projected coordinate $\mathbf{p}_k^{proj}$ can be calculated based on the laser-surface reflection model \cite{li2008single}:
\begin{equation}
    \begin{aligned}
        \mathbf{p}_k^{proj} & = \mathbf{p}_{k}^i - \frac{ \mathbf{v}^c \cdot (\mathbf{p}_{k}^i - \mathbf{p}^o) }{ \mathbf{v}^c \cdot (-\mathbf{v}^c) } (-\mathbf{v}^c)
    \end{aligned}
    \label{fk_3}
\end{equation}

Where the $\mathbf{p}_k^{proj}$ is the projected coordinate of the $k$-th point at the pre-ablation surface, following the negative direction of the laser orientation. The projected distance $s_k^i$ is defined as the point-based distance between the projected coordinate $\mathbf{p}_k^{proj}$ and the laser origin $\mathbf{p}^o$: 
\begin{equation}
    \begin{aligned}
        s_k^i & = || \mathbf{p}_k^{proj} - \mathbf{p}^o ||_2
    \end{aligned}
    \label{fk_4}
\end{equation}
Where $||\cdot||_2$ refers to the L2-norm. In summary, Equation.~\ref{fk_1} to Equation.~\ref{fk_4} describe the FK model that can map a laser incident configuration ($\mathbf{p}^c$, $\mathbf{v}^c$) and a pre-ablation point $\mathbf{p}^i_k$ to a new positioning output $\mathbf{q}_k^i$.

\begin{figure*}[h]
\centering
\includegraphics[scale = 0.42]{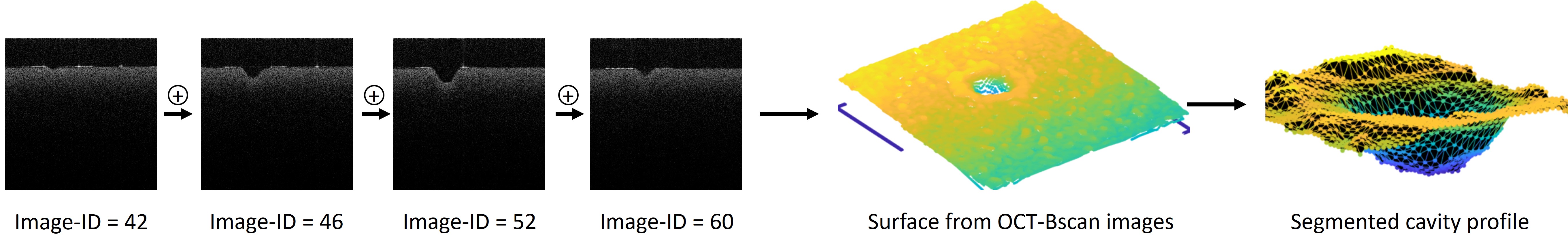}
\caption{Workflow of the cavity reconstruction: The OCT B-scan images (Image-IDs are selected as examples) are collected and segmented with the superficial layers (high contrast between tissue and the air). A surface point cloud is post-processed (e.g. outlier removal and smoothing by 3D interpolation) and the surface cavity region is segmented for analysis.}
\label{fig_oct_bscan_to_3d}
\end{figure*}

\noindent
\textbf{Model fitting for SLP:} Based on Equation.~\ref{fk_2}, we need to estimate the parameters of $f_{SLP}(\cdot)$ in the kinematic model. This is an 1D regression problem and can be formulated as: 
\begin{equation}
    \begin{aligned}
        \min_{\Theta} & \quad \mathcal{L}(\Theta) = \frac{ \sum_{k = 1}^N ( f_{SLP}(\Theta, s_k^i) - d_k^i )^2 }{ N }
    \end{aligned}
\end{equation}
where $\mathcal{L}(\Theta)$ is the Mean-square-error (MSE) loss function. We formulate a dataset of $(s_k^i, d_k^i)$ with a training-validation ratio as $8:2$. The validation dataset is used to select the optimal epoch of the trained SLP model. We use the MATLAB $fitnet$ built-in function to implement the SLP and the Levenberg-Marquardt optimization algorithm for the solver \cite{more1978levenberg}. The testing set is kept separated for verification with a training-testing ratio as $6:4$. 

\noindent
\textbf{Inverse kinematics:} We develop an optimization-based IK solver to determine the optimal laser ablation center $\mathbf{p}^c$ and orientation vector $\mathbf{v}^c$ that can minimize the point-based error to a target point. Given a target point $\mathbf{q}_k^* \in \mathbb{R}^3$, the optimization cost function is: 
\begin{equation}
    \begin{aligned}
        \min_{\mathbf{p}^c, \mathbf{v}^c} & \quad \mathcal{C}_k = || \mathbf{q}_k^i (\mathbf{v}^c, \mathbf{p}^c, \mathbf{q}^i_k) - \mathbf{q}_k^* ||_2^2  \\ 
        \textit{s.t.} & \quad \mathbf{A}_1 
        \begin{bmatrix}
        \mathbf{p}^c & \mathbf{v}^c 
        \end{bmatrix}^T  + \mathbf{b}_1 = 0 \\ 
        & \quad \mathbf{A}_2 \mathbf{p}^c + \mathbf{b}_2 \leq 0,  \quad \mathbf{A}_3 \mathbf{v}^c + \mathbf{b}_3 \leq 0 
    \end{aligned}
    \label{eqn_ik}
\end{equation}
where $\mathcal{C}_k \in \mathbb{R}$ denotes the cost function of the pair of $\mathbf{q}_k^i$ and $\mathbf{q}_k^*$ for the $k$-th index of a surface point. We define $A_1 = [0, 0, -1, 0, 0, 0]$ and $b_1 = \alpha$ ($\alpha$ is an arbitrary value) as the equality constraint to make the laser ablation center moving within a fixed region. The equality constraint is described by a reference plane parallel to the XY-plane with a Z-value of $\alpha$, as shown in Fig.~\ref{fig_laser_tissue_geo_config}b. This has an actual application where the robotic laser scalpel can deliver energy on an arbitrary position at the tissue surface. Additionally, the $(\mathbf{A}_2, \mathbf{b}_2)$ and $(\mathbf{A}_3, \mathbf{b}_3)$ encode the inequality constraints of the upper and lower limits for the ablation center and the orientation vector. This can align with the actual scenario where the robotic laser end-effector can have the geometric constraints for manipulation. 

\noindent
\textbf{Analytical gradients:} To solve the IK problem in (\ref{eqn_ik}) with an efficient optimization solver, we provide the analytical gradient $\frac{\partial \mathcal{C}}{\partial \mathbf{x}}: \mathbb{R} \rightarrow \mathbb{R}^3$ with respect to the cost function. It is noted that the $\mathbf{x} \in \mathbb{R}^3$ can be either $\mathbf{p}^c$ or $\mathbf{v}^c$, and we summarize the gradient as:
\begin{equation}
    \begin{aligned}
        \frac{\partial \mathcal{C}}{\partial \mathbf{x}} & = \frac{\partial \mathcal{C} }{\partial \mathbf{(\Delta q)}} (\frac{\partial d_k^i}{\partial \mathbf{x}} \mathbf{v}^c + d_k^i(\cdot) \frac{\partial \mathbf{v}^c}{\partial \mathbf{x}})
    \end{aligned}
\end{equation}
where $\frac{\partial \mathcal{C} }{\partial \mathbf{(\Delta q)}} = 2 * ( \mathbf{q} (\mathbf{v}^c, \mathbf{p}^c, \mathbf{q}^i_k) - \mathbf{q}^* )$. For the $\frac{\partial d_k^i}{\partial \mathbf{x}}$:
\begin{equation}
    \begin{aligned}
       \frac{\partial d_k^i}{\partial \mathbf{x^c}} & =  \underbrace{ \frac{\partial f_{SLP}}{\partial s_k^i}    }_\text{gradient of SLP}   \frac{\partial s_k^i }{\partial \mathbf{p}_k^{proj} } \frac{\partial \mathbf{p}_k^{proj} }{\partial \mathbf{x}^c }
    \end{aligned}
    \label{gradient_pc}
\end{equation}
where $\frac{\partial d_k^i}{\partial \mathbf{x}^c}$ is a $3 \times 1$ gradient vector. The $\frac{\partial f_{SLP}}{\partial s_k^i}$ denotes the derivative of the output with respect to the input in the SLP. For $\frac{\partial \mathbf{v}^c}{\partial \mathbf{x}}$, we have $\frac{\partial \mathbf{v}^c}{\partial \mathbf{x}} = \mathbf{0}_{3 \times 3}$ if $\mathbf{x} = \mathbf{p}^c$ and $\frac{\partial \mathbf{v}^c}{\partial \mathbf{x}} = \mathbf{I}_{3 \times 3}$ if $\mathbf{x} = \mathbf{v}^c$. The $\frac{\partial s_k^i }{\partial \mathbf{p}_k^{proj} }$ and $\frac{\partial \mathbf{p}_k^{proj} }{\partial \mathbf{x}^c }$ can be calculated based on Equation.~\ref{fk_3} and Equation.~\ref{fk_4}.   

The SLP shows a good mathematical property as a differentiable function to improve the optimization. With the analytical gradients, the problem in (\ref{eqn_ik}) can be solved by the MATLAB Interior point optimization solver \cite{byrd1999interior}. 

\subsection{Optimization-based Robotic Laser Planning Problem}

Given an intraoperative measurement from a 3D sensor, such as OCT, a user (i.e. surgeon) can select a target region-of-interest (ROI) intraoperatively and create a surface target, e.g. a tumor boundary and a tissue cutting ROI. A robotic laser planner can calculate the optimal laser incident configuration to create a simulated post-ablation cavity. This predicted cavity can be applied in surgical planning for fully-automated robotic laser surgery or helping surgeons in evaluating the quality of the current one-shot laser ablation. 

As the robotic laser planning problem is equivalent to the minimization of the alignment offset between the predicted cavity and the surface target, we can formulate the optimization problem as least-squares of point-based errors between the two surfaces:
\begin{equation}
    \begin{aligned}
        \min_{\mathbf{p}^c, \mathbf{v}^c} &  \quad \mathcal{F} = \sum_{k=1}^M \mathcal{C}_k(\mathbf{p}^c, \mathbf{v}^c, \mathbf{p}_k^i, \mathbf{q_k^*}) \\
        \textit{s.t.}  & \quad \mathbf{A}_1 
        \begin{bmatrix}
        \mathbf{p}^c &  \mathbf{v}^c \\
        \end{bmatrix}^T + \mathbf{b}_1 = 0 \\
        & \quad \mathbf{A}_2 \mathbf{p}^c + \mathbf{b}_2 \leq 0, \quad \mathbf{A}_3 \mathbf{v}^c + \mathbf{b}_3 \leq 0 \\
    \end{aligned}
    \label{opt_surface_align}
\end{equation}

where $\mathcal{F}$ is the cost function with respect to $\mathbf{p}^c$ and $\mathbf{v}^c$. The $(\mathbf{A}_1, \mathbf{b}_1)$, $(\mathbf{A}_2, \mathbf{b}_2)$ and $(\mathbf{A}_3, \mathbf{b}_3)$ are the same as the formulation in (\ref{eqn_ik}). The $\mathcal{C}_k(\cdot)$ refers to the same cost function in the IK problem. The $\{\mathbf{p}_k^i\}_{k=1}^M$ and $\{\mathbf{q}_k^*\}_{k=1}^M$ formulate the pair correspondences between the predicted cavity and the surface target. As a proof of concept, this study assumes the point correspondences are given, which is similar to the surface registration problem in \cite{Besl1992}. The same computational framework can be applied to solve more generalized problems when the point-correspondences are not given. 

\subsection{Dataset Collection and Processing for SLP} 

The core problem of the data-driven method is to estimate the parameter of $f_{SLP}$ by giving the training data of the distance-to-laser-center $s_k^i$ and the depth-of-cut $d_k^i$. To create a dataset with multiple cavities, we use the existing robotic laser system \cite{hill2016tumorcnc}, referred as ``TumorCNC", to create various laser orientations and ablation centers on the phantom based on a 10 Watt CO2 laser (Synrad Inc., W.A., United States). Each laser incident configuration can generate a unique surface cavity and this surface profile can be post-processed to obtain a fixed number of data tuples ($s_k^i, d_k^i$).

\subsubsection{Data collection workflow}

\begin{figure}[h]
\centering
\includegraphics[scale = 0.36]{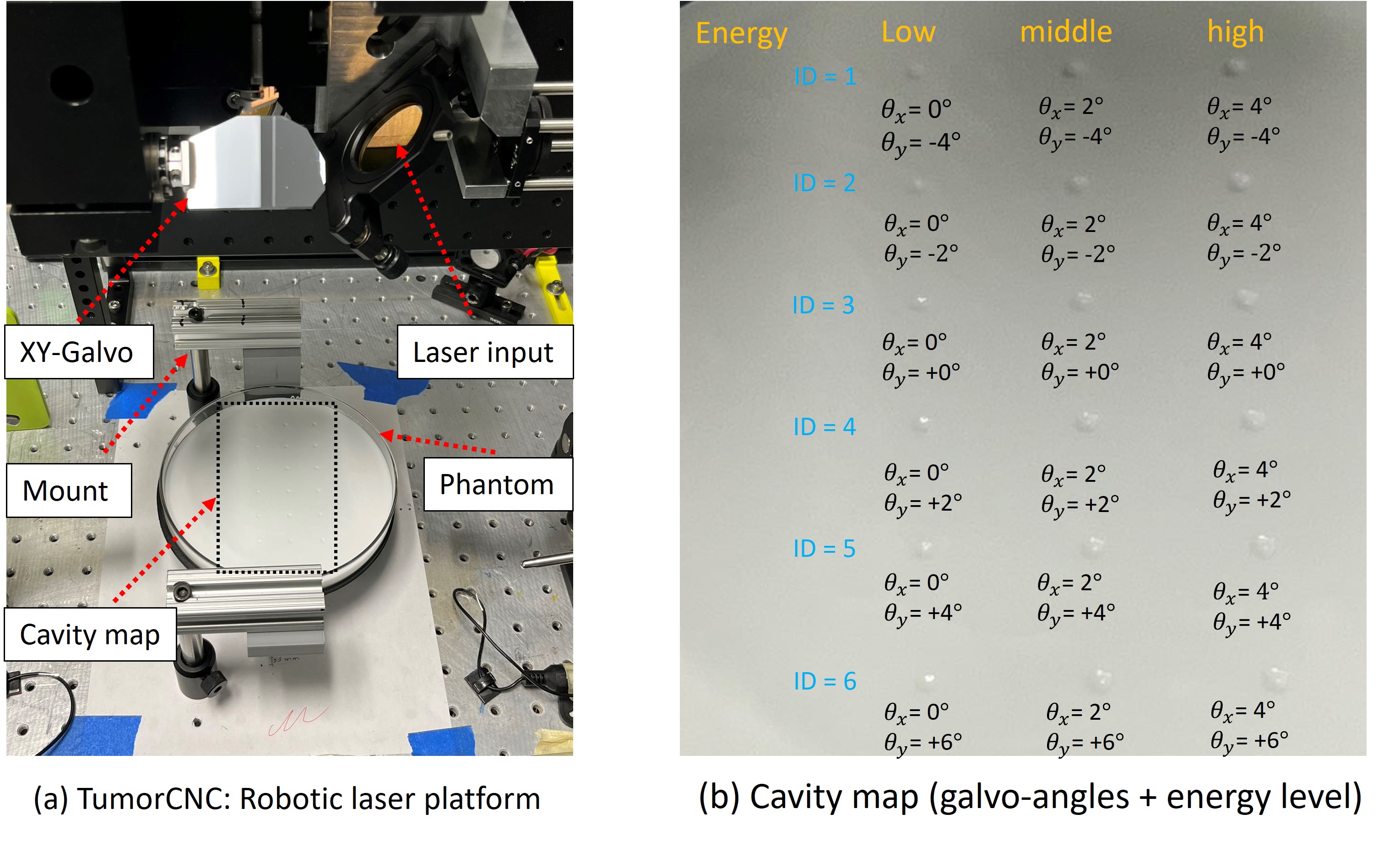}
\caption{The system platform to achieve one-shot laser cavity ablation (a) and the global information of the cavity map (b).}
\label{fig_map_index_config}
\end{figure}

As the proposed data-driven method is laser-and-tissue agnostic, we collect the datasets of three different phantom tissue properties and each with three various energy models. The tissue property is defined by adjusting the ratio of the agarose as $0.5\%$, $1.5\%$, and $2.5\%$ over the total volume of 80 $ml$ phantom, while keeping the ratio of intralipid as $10\%$ of the total phantom volume. The design of phantom has been validated in our prior work \cite{tucker2021creation} for laser ablation. Each phantom tissue is associated with three different energy models classified as 1) low: 25 $J$ (joule) power 2) middle: 56 $J$ and 3) high: 95 $J$, by keeping the laser setting fixed for each ablation. For each energy model, we assign six cavities locations with unique laser incident orientations controlled by the Two-axis galvo mirrors. Thus, each cavity can be considered as a unique data point. The system platform and the layout of the cavity index are shown in Fig.~\ref{fig_map_index_config}a and Fig.~\ref{fig_map_index_config}b.

\begin{figure*}[h]
\centering
\includegraphics[scale = 0.60]{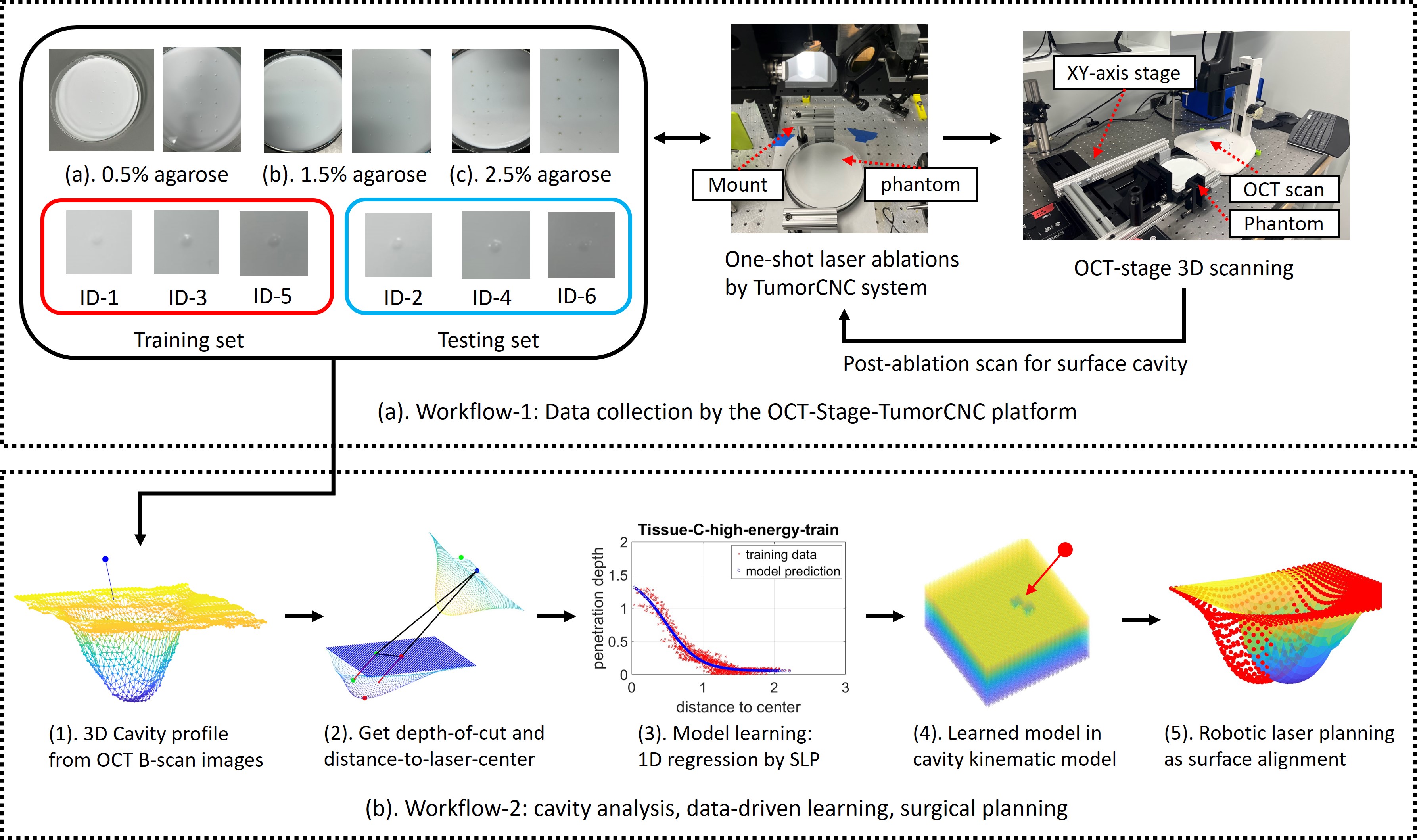}
\caption{Data collection and analysis workflows. (a). The TumorCNC system achieve one-shot laser ablations with various galvo-angles and 18 cavities are obtained. (b). Steps (1) to (3) show that each cavity is converted to multiple data points for data-driven modelling. Steps (4) to (5) show the application of using the learned model for robotic laser planning.}
\label{fig_data_workflow}
\end{figure*}

\subsubsection{Cavity reconstruction from OCT B-scan}

We developed an automated pipeline to attach the post-ablation phantom on the translation system with two 300-mm stepper motorized stages (Thorblab Inc., N.J., USA). The stage visited the pre-defined positions aligned with the cavity ROI, and the Optical coherence tomography (OCT) imaging system (Lumedica, N.C., USA) was applied to scan the surface. The OCT B-scan cross-sectional images were post-processed to 1) apply the bilateral Filter ($\sigma_{color} = 300$, $\sigma_{space} = 150$) with the OpenCV software library to remove the noise and smooth the image, 2) determine the tissue-surface boundary layer based on the selected pixels with the highest intensity in each column, 3) concatenate the segmented boundary to formulate a surface point cloud and 4) manually define the ROI to segment the cavity surface. The workflow is shown in Fig.~\ref{fig_oct_bscan_to_3d}. 

\subsubsection{Post-cavity processing} 

Each cavity used for building the training dataset is post-processed to obtain the data tuples of distance-to-laser-centers and depth-of-cuts. This requires the geometric information including: 1) laser incident orientation, 2) laser incident center, 3) local surface normal orientation and 4) local surface center. The laser incident orientations are given from the calibrated TumorCNC sensor model \cite{hill2016tumorcnc, ma2021stereocnc} based on the galvo-angles. Since the greatest laser intensity matches the maximum depth-of-cut, the laser incident center aligns with the average centers of the data points associated with the top 15\% depth-of-cuts. These data points denote the projected coordinates from the surface points to the laser incident plane following Equation.~\ref{fk_3}. 

The local surface center and normal vectors are estimated based on the surface point cloud with the cavity segmented. Given the geometric information, the depth-of-cuts are equivalent to the lengths of beams by projecting the cavity surface points to the reference plane formulated by the local surface center and normal vector (Fig.~\ref{fig_laser_tissue_geo_config}b and Fig.~\ref{fig_laser_tissue_geo_config}c):
\begin{equation}
    \begin{aligned}
        d_k^i & = ||  \mathbf{p}_{k}^i - \mathbf{ \overline{p} }_k^{proj} ||_2 = || \frac{ (-\mathbf{v}^N) \cdot (\mathbf{p}_{k}^i - \mathbf{p}^N) }{ (-\mathbf{v}^N) \cdot (-\mathbf{v}^c) } (-\mathbf{v}^c) ||_2
    \end{aligned}
    \label{eqn_doc}
\end{equation}
where $\mathbf{ \overline{p} }_k^{proj}$ is the projected coordinate at the local surface. The depth-of-cuts are also matched with the projected coordinates from the cavity surface to the laser incident plane, and thus the distance-to-laser-centers $s_k^i$ can be calculated from Equation.~\ref{fk_4}.

\section{EXPERIMENTS and RESULTS}

\begin{table*}[]
\centering
\renewcommand{\arraystretch}{1.5}
\caption{Summary of regression errors and 3D volumetric errors from the testing dataset}
\label{tbl_error_analysis}
\resizebox{2.0\columnwidth}{!}{%
\begin{tabular}{|c|cccccc|ccccccccc|}
\hline
                                                                                                             & \multicolumn{6}{c|}{Regression: testing error (decimal point = 0.0001)}                                                                                                                                                                                                                                                                                                                                                                       & \multicolumn{9}{c|}{3D volumetric error (decimal point = 0.01)}                                                                                                                                                                                                          \\ \hline
\multirow{2}{*}{\begin{tabular}[c]{@{}c@{}}Phantom-Energy = \\ agarose ratio + \\ energy model\end{tabular}} & \multicolumn{2}{c|}{Num-hidden-layer = 1}                                                                                                            & \multicolumn{2}{c|}{Num-hidden-layer = 3}                                                                                                            & \multicolumn{2}{c|}{Num-hidden-layer = 6}                                                                                       & \multicolumn{3}{c|}{Under-cutting (\%)}                                              & \multicolumn{3}{c|}{Over-cutting (\%)}                                               & \multicolumn{3}{c|}{3D-cavity-IoU (\%)}                                                    \\ \cline{2-16} 
                                                                                                             & \multicolumn{1}{c|}{\begin{tabular}[c]{@{}c@{}}RMSE \\ (mm)\end{tabular}} & \multicolumn{1}{c|}{\begin{tabular}[c]{@{}c@{}}MAE \\ (mm)\end{tabular}} & \multicolumn{1}{c|}{\begin{tabular}[c]{@{}c@{}}RMSE \\ (mm)\end{tabular}} & \multicolumn{1}{c|}{\begin{tabular}[c]{@{}c@{}}MAE \\ (mm)\end{tabular}} & \multicolumn{1}{c|}{\begin{tabular}[c]{@{}c@{}}RMSE \\ (mm)\end{tabular}} & \begin{tabular}[c]{@{}c@{}}MAE \\ (mm)\end{tabular} & \multicolumn{1}{c|}{idx-2} & \multicolumn{1}{c|}{idx-4} & \multicolumn{1}{c|}{idx-6} & \multicolumn{1}{c|}{idx-2} & \multicolumn{1}{c|}{idx-4} & \multicolumn{1}{c|}{idx-6} & \multicolumn{1}{c|}{idx-2}          & \multicolumn{1}{c|}{idx-4}          & idx-6          \\ \hline
0.5\% + low                                                                                                  & \multicolumn{1}{c|}{0.0317}                                               & \multicolumn{1}{c|}{0.0259}                                              & \multicolumn{1}{c|}{0.0311}                                               & \multicolumn{1}{c|}{0.0252}                                              & \multicolumn{1}{c|}{0.0309}                                               & 0.0250                                              & \multicolumn{1}{c|}{8.15}  & \multicolumn{1}{c|}{9.21}  & \multicolumn{1}{c|}{23.88} & \multicolumn{1}{c|}{26.70} & \multicolumn{1}{c|}{15.80} & \multicolumn{1}{c|}{6.81}  & \multicolumn{1}{c|}{\textbf{84.05}} & \multicolumn{1}{c|}{\textbf{87.90}} & \textbf{83.23} \\ \hline
0.5\% + mid                                                                                                  & \multicolumn{1}{c|}{0.0387}                                               & \multicolumn{1}{c|}{0.0317}                                              & \multicolumn{1}{c|}{0.0375}                                               & \multicolumn{1}{c|}{0.0305}                                              & \multicolumn{1}{c|}{0.0372}                                               & 0.0300                                              & \multicolumn{1}{c|}{2.69}  & \multicolumn{1}{c|}{13.69} & \multicolumn{1}{c|}{9.85}  & \multicolumn{1}{c|}{9.94}  & \multicolumn{1}{c|}{0.84}  & \multicolumn{1}{c|}{3.04}  & \multicolumn{1}{c|}{\textbf{93.90}} & \multicolumn{1}{c|}{\textbf{92.24}} & \textbf{93.33} \\ \hline
0.5\% + high                                                                                                 & \multicolumn{1}{c|}{0.0867}                                               & \multicolumn{1}{c|}{0.0662}                                              & \multicolumn{1}{c|}{0.0862}                                               & \multicolumn{1}{c|}{0.0650}                                              & \multicolumn{1}{c|}{0.0863}                                               & 0.0651                                              & \multicolumn{1}{c|}{3.05}  & \multicolumn{1}{c|}{19.96} & \multicolumn{1}{c|}{18.54} & \multicolumn{1}{c|}{7.55}  & \multicolumn{1}{c|}{0.39}  & \multicolumn{1}{c|}{1.94}  & \multicolumn{1}{c|}{\textbf{94.82}} & \multicolumn{1}{c|}{\textbf{88.72}} & \textbf{88.84} \\ \hline
1.5\% + low                                                                                                  & \multicolumn{1}{c|}{0.0308}                                               & \multicolumn{1}{c|}{0.0254}                                              & \multicolumn{1}{c|}{0.0307}                                               & \multicolumn{1}{c|}{0.0253}                                              & \multicolumn{1}{c|}{0.0305}                                               & 0.0250                                              & \multicolumn{1}{c|}{13.37} & \multicolumn{1}{c|}{4.83}  & \multicolumn{1}{c|}{8.13}  & \multicolumn{1}{c|}{4.84}  & \multicolumn{1}{c|}{17.94} & \multicolumn{1}{c|}{10.20} & \multicolumn{1}{c|}{\textbf{90.49}} & \multicolumn{1}{c|}{\textbf{89.31}} & \textbf{90.93} \\ \hline
1.5\% + mid                                                                                                  & \multicolumn{1}{c|}{0.0560}                                               & \multicolumn{1}{c|}{0.0442}                                              & \multicolumn{1}{c|}{0.0553}                                               & \multicolumn{1}{c|}{0.0431}                                              & \multicolumn{1}{c|}{0.0554}                                               & 0.0431                                              & \multicolumn{1}{c|}{16.57} & \multicolumn{1}{c|}{2.15}  & \multicolumn{1}{c|}{3.42}  & \multicolumn{1}{c|}{0.89}  & \multicolumn{1}{c|}{10.67} & \multicolumn{1}{c|}{14.31} & \multicolumn{1}{c|}{\textbf{90.53}} & \multicolumn{1}{c|}{\textbf{93.85}} & \textbf{91.59} \\ \hline
1.5\% + high                                                                                                 & \multicolumn{1}{c|}{0.0640}                                               & \multicolumn{1}{c|}{0.0499}                                              & \multicolumn{1}{c|}{0.0631}                                               & \multicolumn{1}{c|}{0.0494}                                              & \multicolumn{1}{c|}{0.0633}                                               & 0.0493                                              & \multicolumn{1}{c|}{6.40}  & \multicolumn{1}{c|}{2.92}  & \multicolumn{1}{c|}{10.79} & \multicolumn{1}{c|}{3.20}  & \multicolumn{1}{c|}{6.83}  & \multicolumn{1}{c|}{6.27}  & \multicolumn{1}{c|}{\textbf{95.12}} & \multicolumn{1}{c|}{\textbf{95.22}} & \textbf{91.27} \\ \hline
2.5\% + low                                                                                                  & \multicolumn{1}{c|}{0.0374}                                               & \multicolumn{1}{c|}{0.0287}                                              & \multicolumn{1}{c|}{0.0368}                                               & \multicolumn{1}{c|}{0.0281}                                              & \multicolumn{1}{c|}{0.0357}                                               & 0.0272                                              & \multicolumn{1}{c|}{14.27} & \multicolumn{1}{c|}{14.09} & \multicolumn{1}{c|}{11.00} & \multicolumn{1}{c|}{6.38}  & \multicolumn{1}{c|}{4.65}  & \multicolumn{1}{c|}{6.96}  & \multicolumn{1}{c|}{\textbf{89.25}} & \multicolumn{1}{c|}{\textbf{90.16}} & \textbf{90.83} \\ \hline
2.5\% + mid                                                                                                  & \multicolumn{1}{c|}{0.0473}                                               & \multicolumn{1}{c|}{0.0366}                                              & \multicolumn{1}{c|}{0.0430}                                               & \multicolumn{1}{c|}{0.0329}                                              & \multicolumn{1}{c|}{0.0427}                                               & 0.0329                                              & \multicolumn{1}{c|}{7.06}  & \multicolumn{1}{c|}{5.90}  & \multicolumn{1}{c|}{7.63}  & \multicolumn{1}{c|}{6.19}  & \multicolumn{1}{c|}{7.37}  & \multicolumn{1}{c|}{3.81}  & \multicolumn{1}{c|}{\textbf{93.34}} & \multicolumn{1}{c|}{\textbf{93.42}} & \textbf{94.17} \\ \hline
2.5\% + high                                                                                                 & \multicolumn{1}{c|}{0.0813}                                               & \multicolumn{1}{c|}{0.0589}                                              & \multicolumn{1}{c|}{0.0779}                                               & \multicolumn{1}{c|}{0.0543}                                              & \multicolumn{1}{c|}{0.0769}                                               & 0.0524                                              & \multicolumn{1}{c|}{10.37} & \multicolumn{1}{c|}{9.40}  & \multicolumn{1}{c|}{12.95} & \multicolumn{1}{c|}{3.94}  & \multicolumn{1}{c|}{4.65}  & \multicolumn{1}{c|}{5.78}  & \multicolumn{1}{c|}{\textbf{92.61}} & \multicolumn{1}{c|}{\textbf{92.81}} & \textbf{90.29} \\ \hline
Average                                                                                                      & \multicolumn{1}{c|}{0.0527}                                               & \multicolumn{1}{c|}{0.0408}                                              & \multicolumn{1}{c|}{0.0513}                                               & \multicolumn{1}{c|}{0.0393}                                              & \multicolumn{1}{c|}{0.0510}                                               & 0.0389                                              & \multicolumn{1}{c|}{9.10}  & \multicolumn{1}{c|}{9.13}  & \multicolumn{1}{c|}{11.80} & \multicolumn{1}{c|}{7.74}  & \multicolumn{1}{c|}{7.68}  & \multicolumn{1}{c|}{6.57}  & \multicolumn{1}{c|}{\textbf{91.57}} & \multicolumn{1}{c|}{\textbf{91.51}} & \textbf{90.50} \\ \hline
Standard deviation                                                                                           & \multicolumn{1}{c|}{0.0209}                                               & \multicolumn{1}{c|}{0.0149}                                              & \multicolumn{1}{c|}{0.0206}                                               & \multicolumn{1}{c|}{0.0143}                                              & \multicolumn{1}{c|}{0.0206}                                               & 0.0143                                              & \multicolumn{1}{c|}{4.90}  & \multicolumn{1}{c|}{5.91}  & \multicolumn{1}{c|}{6.12}  & \multicolumn{1}{c|}{7.57}  & \multicolumn{1}{c|}{6.12}  & \multicolumn{1}{c|}{3.80}  & \multicolumn{1}{c|}{\textbf{3.48}}  & \multicolumn{1}{c|}{\textbf{2.56}}  & \textbf{3.15}  \\ \hline
\end{tabular}%
}
\end{table*}

\subsection{Function Fitting for SLP}

\subsubsection{Dataset preparation} 

Each phantom-energy model included 6 surface cavities. It is noted that the discrete data points for training the SLP were extracted from each cavity, i.e. each cavity can contain a lot of data points. We separated six cavities as training and testing datasets with a 50$\%$ ratio (3 for training and 3 for testing). This is different than the common training-testing splitting factor since this study requires more data to test the feasibility of fitting the function. For each column of the data points, the index-1, index-3, and index-5 were used for training and index-2, index-4, and index-6 were used for the testing dataset. The three cavities used for training were post-processed to data points encoded with $(s_k^i, d_k^i)$. The points for each cavity were sampled based on the uniformly distributed random numbers. A total number of 1980 data points (each cavity has 660 points) was used to formulate the training and validation dataset. The testing dataset has 1320 data points. 

\subsubsection{SLP and MLP comparison} 

To validate the feasibility of using SLP, we compared the SLP with the multilayer perceptron (MLP) network with numbers of hidden layers as 1, 3 and 6 respectively. The regression analysis of the testing dataset is shown in Table.~\ref{tbl_error_analysis}. The root-mean-square-error (RMSE) and the mean absolute error (MAE) were applied to measure the performance of the SLP.

\subsection{3D Volumetric Error Analysis}

The regression analysis cannot report the 3D volumetric offset between the predicted cavity and the reference ground truth. Therefore, we introduce a 3D volumetric error analysis to evaluate the difference between the predicted and the reference cavity. Given the same ablation center and the laser incident vector, the data-driven cavity prediction model was applied to create a predicted cavity. The offset was measured between the predicted cavity profiles and the ground truth information from the testing cavities (testing cavity index as 2, 4, 6). The ratios were reported for the cases of under-cutting, over-cutting and 3D-cavity-intersection-of-union (3D-cavity-IoU). The under-cutting ratio is defined as the ratio of volumetric tissue that should be cut but is not being cut over the reference cavity volume as ground truth. The over-cutting ratio refers to the ratio of volumetric tissue that should not be cut but is being cut over the same reference cavity volume. The 3D-cavity-IoU denotes the ratio of volume overlapping between predicted cavity and the ground truth.

\subsubsection{3D volume calculation}

\begin{figure}[h]
\centering
\includegraphics[scale = 0.35]{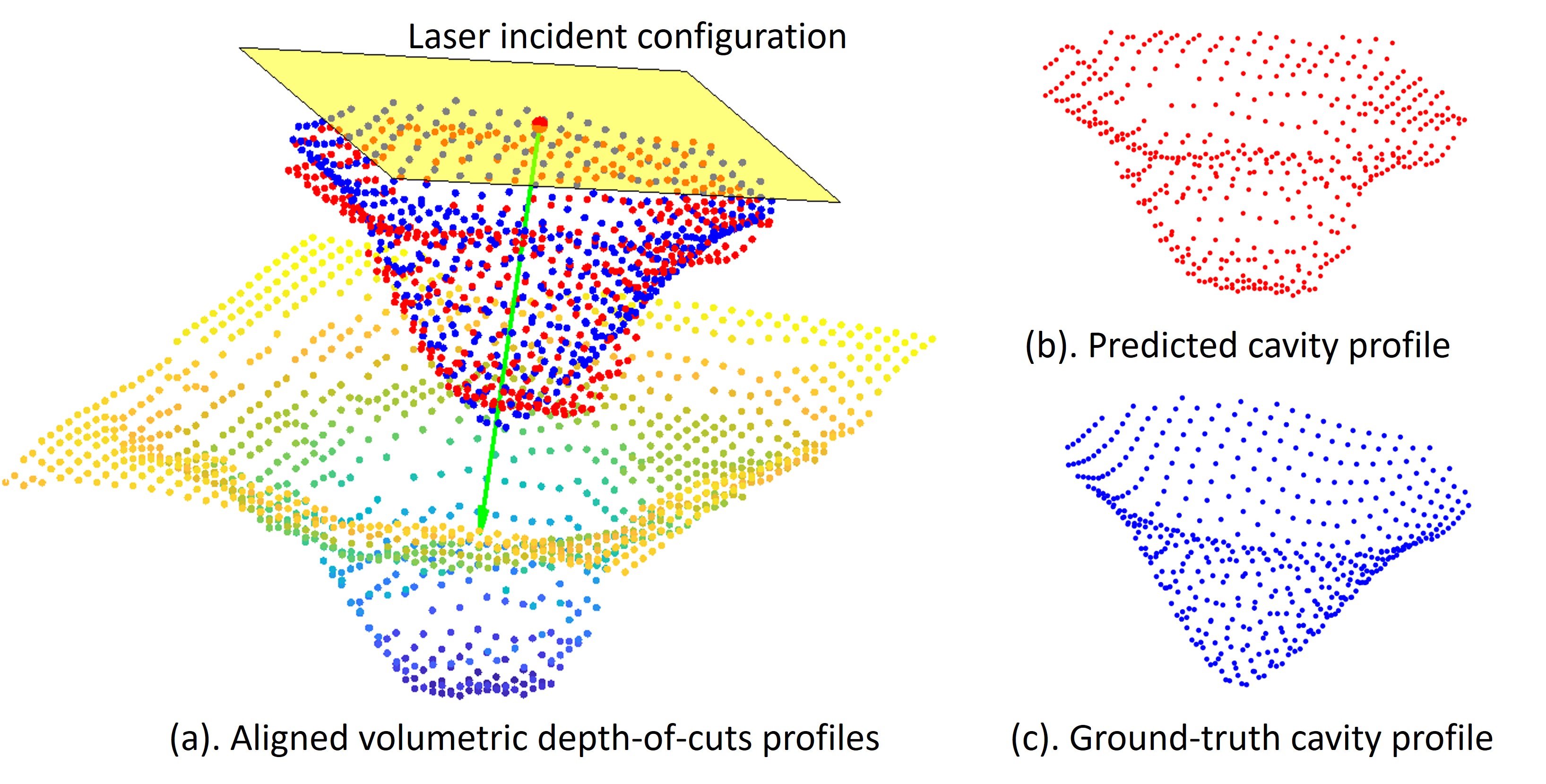}
\caption{3D graphical visualization of volumes between the predicted cavity and the ground truth. The volumes are mapped to the same ROI at the laser incident plane.}
\label{fig_3d_vol_error}
\end{figure}

On the laser incident plane (Fig.~\ref{fig_laser_tissue_geo_config}c), a region-of-interest (ROI) was defined with a radius as 1.0 mm (2.0 mm diameter covers the range of the laser beam spot size). A fixed number of points were uniformly sampled in the ROI, and each point corresponds to a unique depth-of-cut from the predicted cavity and the ground truth (referred as ``gt"). The volume of the cavity was computed as the integral of the depth-of-cuts in the ROI: 
\begin{equation}
\begin{aligned}
    V & = \int_{\overline{\mathbf{p}} \in ROI} d V(\overline{\mathbf{p}}) = \int_{\overline{\mathbf{p}} \in ROI} D( \overline{\mathbf{p}} ) ~ d A( \overline{\mathbf{p}} ) 
    \label{3d_vol_eqn}
\end{aligned}
\end{equation}
Where $d V(\overline{\mathbf{p}})$ denotes a very small 3D volume. The $D( \overline{\mathbf{p}})$ (capital $D$ to differentiate the differential symbol $d$) is the depth-of-cut associated with the projected coordinate $\overline{\mathbf{p}}$ in the ROI at the laser incident plane. The $d A(\overline{\mathbf{p}}) $ represents a very small area aligned with $D( \overline{\mathbf{p}} )$. The graphical visualization of the over-cut, the under-cut and the two caviters are shown in Fig.~\ref{fig_3d_vol_error}.

\begin{table*}[]
\centering
\caption{ Success rate of simulated robotic laser planning experiments (each model has 225 test cases).}
\label{tbl_align_error}
\resizebox{2.0\columnwidth}{!}{%
\def\arraystretch{1.2}%
\begin{tabular}{|c|c|c|c|c|c|c|c|c|c|}
\hline
  Model      & 0.5\% + low & 0.5\% + mid & 0.5\% + high & 1.5\% + low & 1.5\% + mid & 1.5\% + high & 2.5\% + low & 2.5\% + mid & 2.5\% + high \\ \hline
Success rate (\%) & 100.00      & 99.11       & 100.00       & 91.11       & 99.56      & 99.56        & 94.67       & 98.22       & 99.56      \\ \hline
\end{tabular}%
}
\end{table*}

\subsubsection{Calculation of over-cut, under-cut and 3D-cavity-IoU} 

For each $\overline{\mathbf{p}}$, $D(\overline{\mathbf{p}})^{predict} \geq D(\overline{\mathbf{p}})^{gt}$ denotes the set of the over-cutting data points and $D(\overline{\mathbf{p}})^{predict} < D(\overline{\mathbf{p}})^{gt}$ indicates the under-cutting one. The 3D-cavity-IoU is computed as $\frac{2 * V_{overlap}}{V_{gt} + V_{predict}}$, where $V_{overlap}$ is calculated by taking the integral of data points from the set $\{ D(\overline{\mathbf{p}})^{overlap} \} = \{ D(\overline{\mathbf{p}}) | D(\overline{\mathbf{p}}) = D(\overline{\mathbf{p}})^{predict}, D(\overline{\mathbf{p}})^{predict} \leq D(\overline{\mathbf{p}})^{gt} \}$ $\cap$  $\{ D(\overline{\mathbf{p}}) | D(\overline{\mathbf{p}}) = D(\overline{\mathbf{p}})^{gt}, D(\overline{\mathbf{p}})^{predict} \geq D(\overline{\mathbf{p}})^{gt} \}$. The $V_{gt}$ and $V_{predict}$ are computed based on Equation.~\ref{3d_vol_eqn}. Each testing cavity is uniquely mapped to the ratios of over-cut, under-cut and 3D-cavity-IoU, as shown in Table.~\ref{tbl_error_analysis}.

\subsection{Simulation Experiments for Robotic Laser Planning}

We conducted simulation experiments to 1) solve a simplified robotic laser planning problem modelled as a cavity alignment problem in (\ref{opt_surface_align}), and 2) verify the feasibility of the data-driven laser-tissue cavity prediction model. It is noted that the output of the optimization solver is the 6-dof laser incident configuration. Each laser incident configuration can create a unique surface cavity based on the data-driven kinematic model. This cavity is used as the target to formulate the least-squares objective function. We summarize the simulation experiments as: 

\subsubsection{Sampled laser incident configurations as ground truth} 

A fixed number of laser incident orientations were defined as the ground truth for the optimization. These orientations were calculated by multiplying the rotation matrix with a fixed laser incident vector. The rotation matrix was formulated by the angle vector $(\theta_x, \theta_y, 0)$ sampled from $(\theta_x, \theta_y) \in [-30^{\circ}, +30^{\circ}] \times [-30^{\circ}, +30^{\circ}]$ with an interval of $15^{\circ}$. Each paired of $(\theta_x, \theta_y)$ formulates a unique rotation matrix that can be multiplied by the laser incident vector to create a different laser orientation. In addition, we defined a fixed laser incident center associated with each laser incident orientation. This can formulate a set of randomized 6-dof laser incident configurations with totally 225 combinations. 

\subsubsection{Randomized optimization initialization} 

Each laser incident configuration (ground truth) was perturbed both with the position and orientation for optimization initialization. The initialized positions were added with a vector denoted as $(\sigma_x, \sigma_y, 0)$, where $\sigma_x$ and $\sigma_y$ were uniformly distributed random numbers multiplied by a factor of $0.5$ (unit: mm), i.e. $\sigma_x, \sigma_y \in [-0.5, +0.5]$. The initialized laser orientations were uniformly sampled with XY-angles from $(\theta_x, \theta_y) \in [-15^{\circ}, +15^{\circ}] \times [-15^{\circ}, +15^{\circ}]$, with an interval of $15^{\circ}$. There are totally 9 combinations of the initialized settings. 

\subsubsection{Successful optimization}

The output of the optimization solver is the 6-dof laser incident configuration. The optimal and the ground truth laser incident configurations were defined as $\mathbf{X}_{opt} = (\mathbf{p}_{opt}, \mathbf{v}_{opt}) \in \mathbb{R}^6$ and $\mathbf{X}_{gt} = (\mathbf{p}_{gt}, \mathbf{v}_{gt}) \in \mathbb{R}^6$. A successful optimization is counted when it satisfies $||\mathbf{X}_{opt}-\mathbf{X}_{gt}||_2 \leq 0.00001$. For each phantom-energy model, experiments were conducted for 225 times with various ground truth and initialized settings. The success rates of the simulation experiments are reported in Table.~\ref{tbl_align_error}.

\section{DISCUSSIONS}

\begin{figure}[h]
\centering
\includegraphics[scale = 0.26]{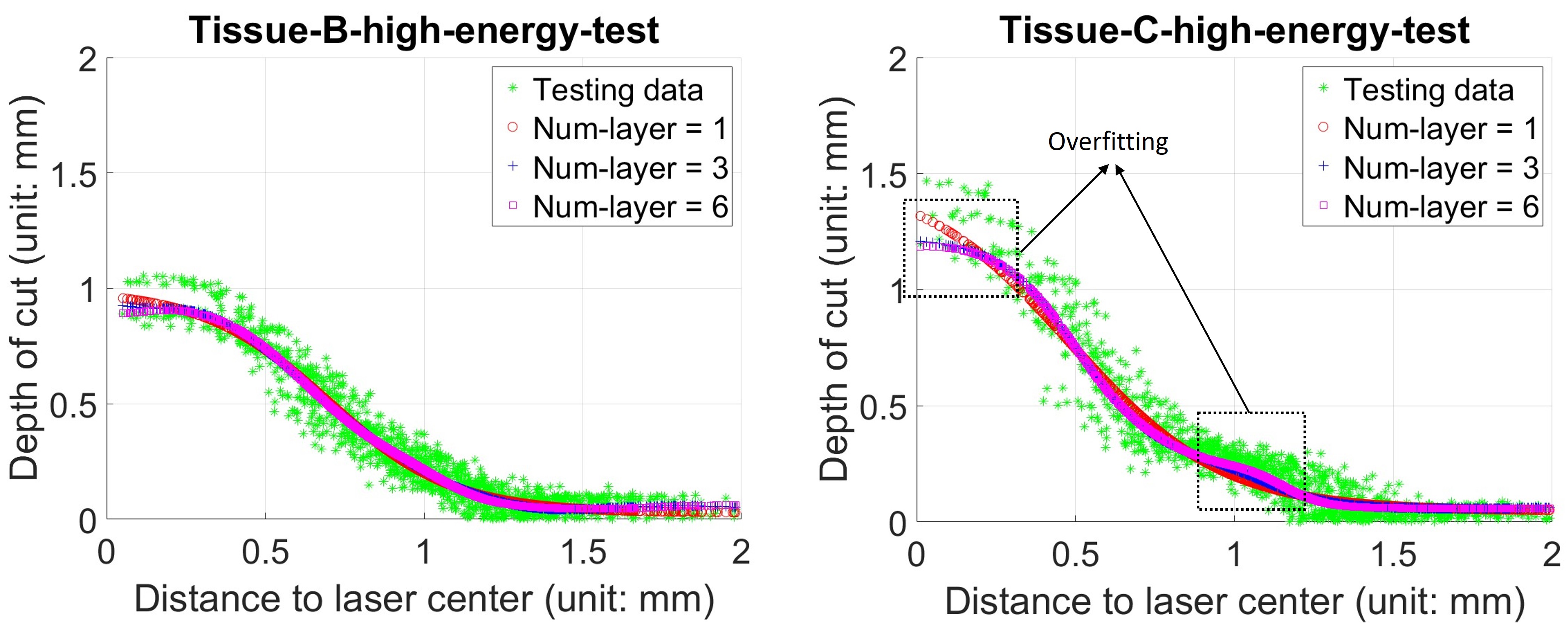}
\caption{ Comparisons of MLPs with various hidden layers (1, 3, 6). The SLP shows better generalization compared with the two MLPs. Tissue-B and Tissue-C denote $1.5\%$-agarose and $2.5\%$-agarose. }
\label{fig_mlp_compare}
\end{figure}

\noindent
\textbf{Analysis of regression error for SLP:} The average RMSE of the different MLPs (variety of 1, 3, 6 hidden layers) are reported as $0.0527 \pm 0.0209$ mm, $0.0513 \pm 0.0206$ mm and $0.0510 \pm 0.0206$ mm respectively. The mean absolute errors (MAE) are summarized as $0.0408 \pm 0.0149$ mm, $0.0393 \pm 0.0143$ mm and $0.0389 \pm 0.0143$ mm. The SLP shows comparable performance with the 3-layers and 6-layers MLPs about the average RMSE and MAE. The 3-layer and 6-layer MLPs show very similar results and it is likely that the MLPs reach to a limit to describe a simple data distribution with more parameters. Additionally, the SLP has better generalization performance with less parameters and thus can minimize the probability of over-fitting, as shown in Fig.~\ref{fig_mlp_compare}. These results can verify the feasibility of using the SLP for the phantom study. However, it does not indicate that other MLPs cannot be applied for the same problem. For some cases, such as ex-vivo tissue studies, where the laser-tissue mechanism is complicated, the MLPs with greater number of hidden layers can be more suitable compared with SLP.

\noindent
\textbf{Analysis of 3d volumetric error:} The over-cutting and under-cutting ratios are comparable with relatively high variances, but they are all smaller than the ratio of $12\%$. This indicates that an one-shot laser ablation can have 12\% of the volume being over-cutting or under-cutting at the worst case. This is not optimal since with a large energy model for a greater size of tissue removal, the 12\% of the over-cut can contribute to a relatively large tissue offset for the surgical planning. The 3D-cavity-IoUs are reported as $91.57 \pm 3.48\%$, $91.51 \pm 2.56\%$ and $90.50 \pm 3.15\%$ for the three testing cavities. This indicates that about $91.1\%$ of the tissue can be precisely removed for an one-shot ablation.

\noindent
\textbf{Analysis of robotic laser planning experiments:} The results of the simulated robotic laser planning experiments show an average of $97.98\%$ success rates. This validates the feasibility of using the analytical gradient for solving the IK problem in (\ref{opt_surface_align}) and the implementation of SLP in the kinematic model. However, there exist some cases that fail to find the optimal solutions during the optimization. This is likely caused by the problem of the in-feasible initial guess for the optimization solver.

\noindent 
\textbf{Limitations and future works:} The surface of the phantom is assumed to be planar but the tissue surface geometry can be arbitrary in actual medical applications. In addition, this study only conducts the simulation experiments. Future works include the experiments with different laser-tissue geometry such as using non-planar surface objects, and the implementation of the methods in an actual robot system.  

\section{Conclusions} 

This work explores a research problem about whether the 3D laser-tissue interaction can be directly learned from the data instead of historical heuristic models. A single-layer perceptron (SLP) network was used to learn the geometric relation between the tissue surface and the laser energy model. The SLP was successfully implemented in a novel cavity prediction model to solve a simplified robotic laser planning problem, which shows potential applications of automated pathological tissue removal and surgical planning. 


\bibliographystyle{IEEEtran}
\bibliography{IEEEabrv, IEEEexample}

\section{Appendix}

\subsection{Derivatives for the Cost Function}

To compute the gradient of the cost function in Equation \ref{eqn_ik}, we provide the derivations for $\frac{f_{SLP}(s_k^i)}{s_k^i}$, $\frac{\partial s_k^i }{\partial \mathbf{p}_k^{proj} }$, $\frac{\partial \mathbf{p}_k^{proj} }{\partial \mathbf{p}^c}$ and $\frac{\partial \mathbf{p}_k^{proj} }{\partial \mathbf{v}^c}$. 

\subsubsection{$\frac{\partial f_{SLP}(s_k^i)}{\partial s_k^i}$} 

The $s_k^i$ is defined as the projected distance at the laser incident plane. The $f_{SLP}(s_k^i)$ represents the end-to-end single layer perceptron network (SLP) to map $s_k^i$ to the depth-of-cut $d_k^i$ based on Equation.~\ref{fk_2}. 

To compute $\frac{\partial f_{SLP}(s_k^i)}{\partial s_k^i}$, a generalized SLP is defined as $y_i = f_{SLP}(x_i)$ where $x_i \in \mathbb{R}$ and $y_i \in \mathbb{R}$ are denoted as the input and the output. The relation between $x_i$ and $y_i$ are the non-linear mapping with a standardized single layer perceptron network architecture: 
\begin{equation}
    \begin{aligned}
        y_i & = f_{SLP}(x_i) = \mathcal{T}_y ( f_{tansig} ( w_1 \mathcal{T}_x (x_i) + b_1 ) * w_2 + b_2 ) 
    \end{aligned}
\end{equation}
Where $(w_1, b_1, w_2, b_2)$ are the weights and bias for the SLP. The $\mathcal{T}_x$ and $\mathcal{T}_y$ represent the given linear min-max transform for the input and output. For $\mathcal{T}_x$, we have: 
\begin{equation}
    \begin{aligned}
         \mathcal{T}_x(x) & = \frac{x - x_{min}^{input}}{x_{max}^{input} - x_{min}^{input}} * (x_{max}^{*} - x_{min}^{*} ) + x_{min}^{*}
    \end{aligned}
\end{equation}
Where $x_{max}^{*} = +1$ and $x_{min}^{*} = -1$. The $x_{max}^{input}$ and $x_{min}^{input}$ are the max and min of the inputs (distance-to-laser-center $s_k^i$) in the training dataset. The goal of $\mathcal{T}_x(x)$ is to normalize the input to the range of $[-1, 1]$. As $x_{max}^{*}$, $x_{min}^{*}$, $x_{max}^{input}$ and $x_{min}^{input}$ are given, the derivative of $\mathcal{T}_x(x)$ can be computed as: 
\begin{equation}
    \begin{aligned}
         \frac{\partial \mathcal{T}_x(x)}{\partial x} & = \frac{ x_{max}^{*} - x_{min}^{*}  } { x_{max}^{input} - x_{min}^{input} }
    \end{aligned}
\end{equation}
For $\mathcal{T}_y$, we have: 
\begin{equation}
    \begin{aligned}
       \mathcal{T}_y(y) & = \frac{y - y_{min}^*}{y_{max}^* - y_{min}^*} ( y_{max}^{label} - y_{min}^{label}) + y_{min}^{label}
    \end{aligned}
\end{equation}
Where $y_{max}^* = 1$, $y_{min}^* = -1$. The $y_{max}^{label}$ and $x_{min}^{label}$ are the max and min of the outputs (depth-of-cuts) in the training dataset. Thus, $\mathcal{T}_y(y)$ can convert the network output to the correct value with the physical interpretation. As $y_{max}^{*}$, $y_{min}^{*}$, $y_{max}^{label}$ and $y_{min}^{label}$ are given, the derivative of $\mathcal{T}_y(y)$ can be computed as: 
\begin{equation}
    \begin{aligned}
         \frac{\partial \mathcal{T}_y(y)}{\partial y} & = \frac{  y_{max}^{label} - y_{min}^{label} }{ y_{max}^* - y_{min}^* }
    \end{aligned}
\end{equation}

Therefore, the $\mathcal{T}_x^{'}$ and $\mathcal{T}_y^{'}$ can be calculated given the training dataset.

For the activation function, the $f_{tansig}(\cdot)$ is the tansig function with the representation as: 
\begin{equation}
    \begin{aligned}
        f_{tansig}(z) & = \frac{e^{z} - e^{-z} }{ e^z + e^{-z} }
    \end{aligned}
\end{equation}
The derivative can be computed as: 
\begin{equation}
    \begin{aligned}
        f^{'}_{tansig}(z) & = 1 - ( \frac{e^{z} - e^{-z} }{ e^z + e^{-z} } ) ^ 2
    \end{aligned}
\end{equation}
In summary, we can compute the derivative of the SLP as: 
\begin{equation}
    \begin{aligned}
        \frac{\partial y_i}{\partial x_i} & = w_1  w_2 \mathcal{T}_y^{'} \mathcal{T}_x^{'}  f^{'}_{tansig}(z) 
    \end{aligned}
\end{equation}
where $z = w_1 \mathcal{T}_x(x_i) + b_1$.  

\subsubsection{$\frac{\partial s_k^i }{\partial \mathbf{p}_k^{proj}}$} 
For $s_k^i = ||\mathbf{p}_k^{proj} - \mathbf{p}^o ||_2$, we have:
\begin{equation}
    \begin{aligned}
        \frac{\partial s_k^i }{\partial \mathbf{p}_k^{proj} } & =  \frac{ \mathbf{p}_k^{proj} - \mathbf{p}^o }{ || \mathbf{p}_k^{proj} - \mathbf{p}^o ||_2 }
    \end{aligned}
\end{equation}

\subsubsection{$\frac{\partial \mathbf{p}_k^{proj} }{\partial \mathbf{p}^c}$} 

Based on Equation.~\ref{fk_3}, we have:

\begin{equation}
\begin{aligned}
    \frac{\partial \mathbf{p}_k^{proj} }{\partial \mathbf{p}^c}  & = \mathbf{I} - \frac{ (-\mathbf{v}^c) ( \mathbf{v}^c )^T }{ \mathbf{v}^c \cdot (-\mathbf{v}^c) }
\end{aligned}
\end{equation}
where $\mathbf{v}^c$ denotes the laser orientation vector.

\subsubsection{$\frac{\partial \mathbf{p}_k^{proj} }{\partial \mathbf{v}^c}$} 

Based on Equation.~\ref{fk_3}, we have:

\begin{equation}
\begin{aligned}
     \frac{\partial \mathbf{p}_k^{proj} }{\partial \mathbf{v}^c}  = & - \frac{ (-\mathbf{v}^c \cdot \mathbf{v}^c) * [\mathbf{v}^c \cdot (\mathbf{p}^{i}_{k} -\mathbf{p}^o)]  } { ((-\mathbf{v}^c) \cdot \mathbf{v}^c)^2 } \mathbf{I}_{3 \times 3}
      \\
      & + 
     \frac{ \mathbf{v}^c \cdot (\mathbf{p}_{k}^i - \mathbf{p}^o) }{ ( (-\mathbf{v}^c) \cdot \mathbf{v}^c)^2 }  (-\mathbf{v}^c) (\mathbf{v}^c)^T  
\end{aligned}
\end{equation}
Where $\mathbf{p}^o$ is the projected center (from $\mathbf{p}^c$) at the laser incident plane. The $\mathbf{p}_{k}^i$ is the pre-defined coordinate at the pre-ablation surface associated with $\mathbf{p}_k^{proj}$. 

\subsection{Regression Analysis}

\begin{figure*}[h]
\centering
\includegraphics[scale = 0.95]{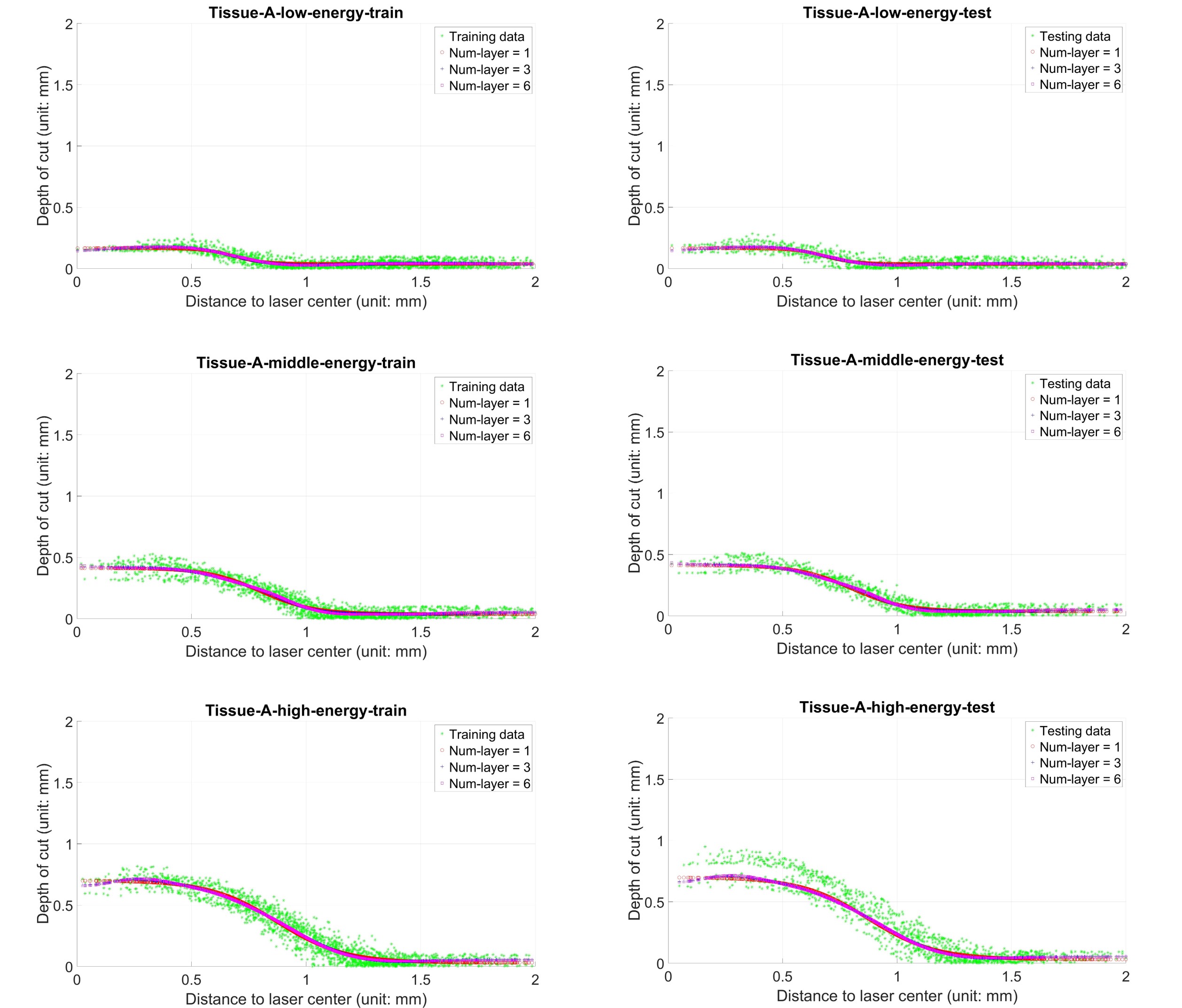}
\caption{Regression analysis of training-testing datasets for Tissue-A ($0.5\%$ agar). The scattered dots depict the data points of the dataset (training as left column and testing as right column). The data labelled with different colors represent the outputs from the neural networks with various layers for regression. }
\label{fig_regression_A}
\end{figure*}

\begin{figure*}[h]
\centering
\includegraphics[scale = 0.95]{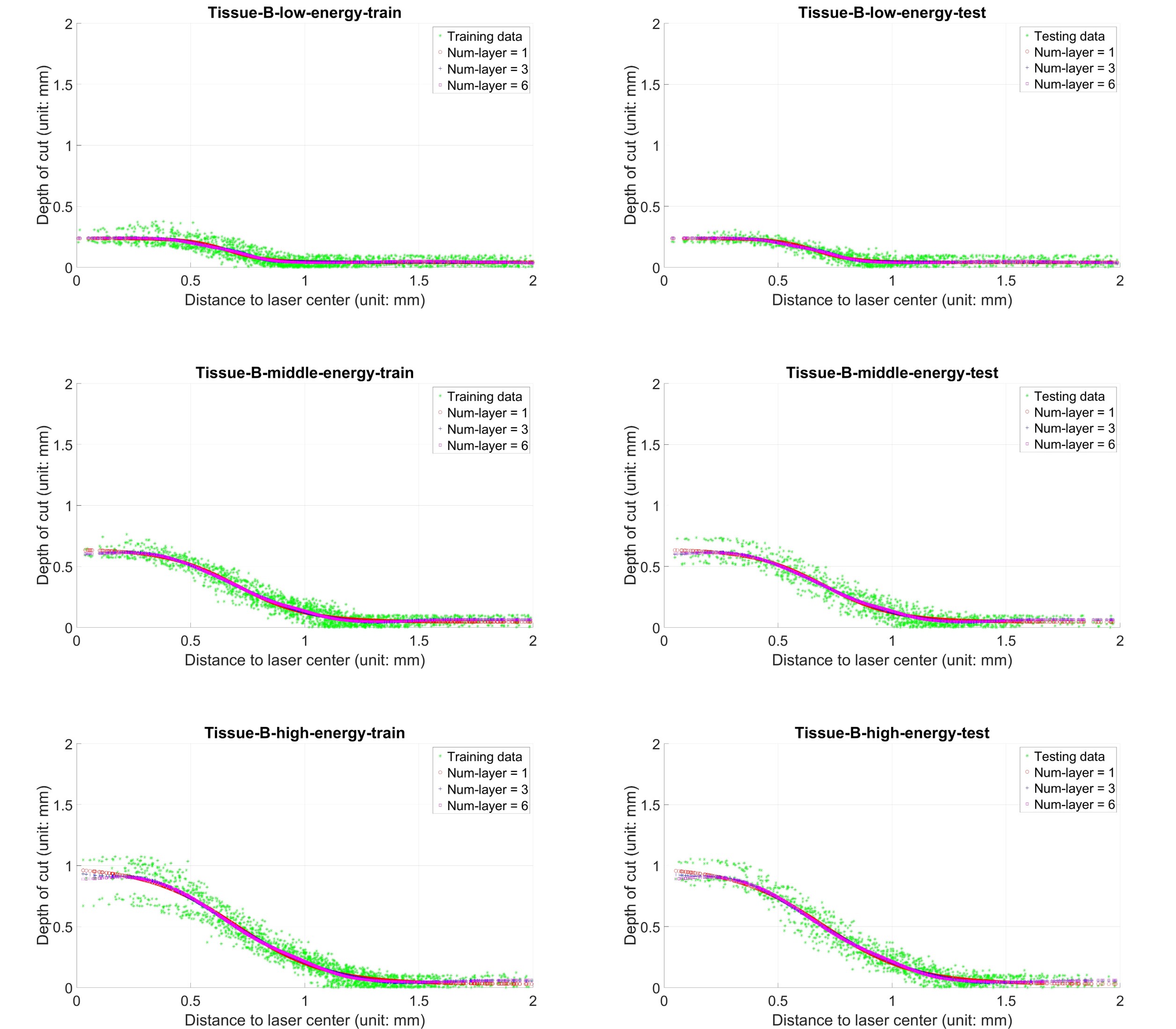}
\caption{Regression analysis of training-testing datasets for Tissue-B ($1.5\%$ agar). The scattered dots depict the data points of the dataset (training as left column and testing as right column). The data labelled with different colors represent the outputs from the neural networks with various layers for regression.}
\label{fig_regression_B}
\end{figure*}

\begin{figure*}[h]
\centering
\includegraphics[scale = 0.95]{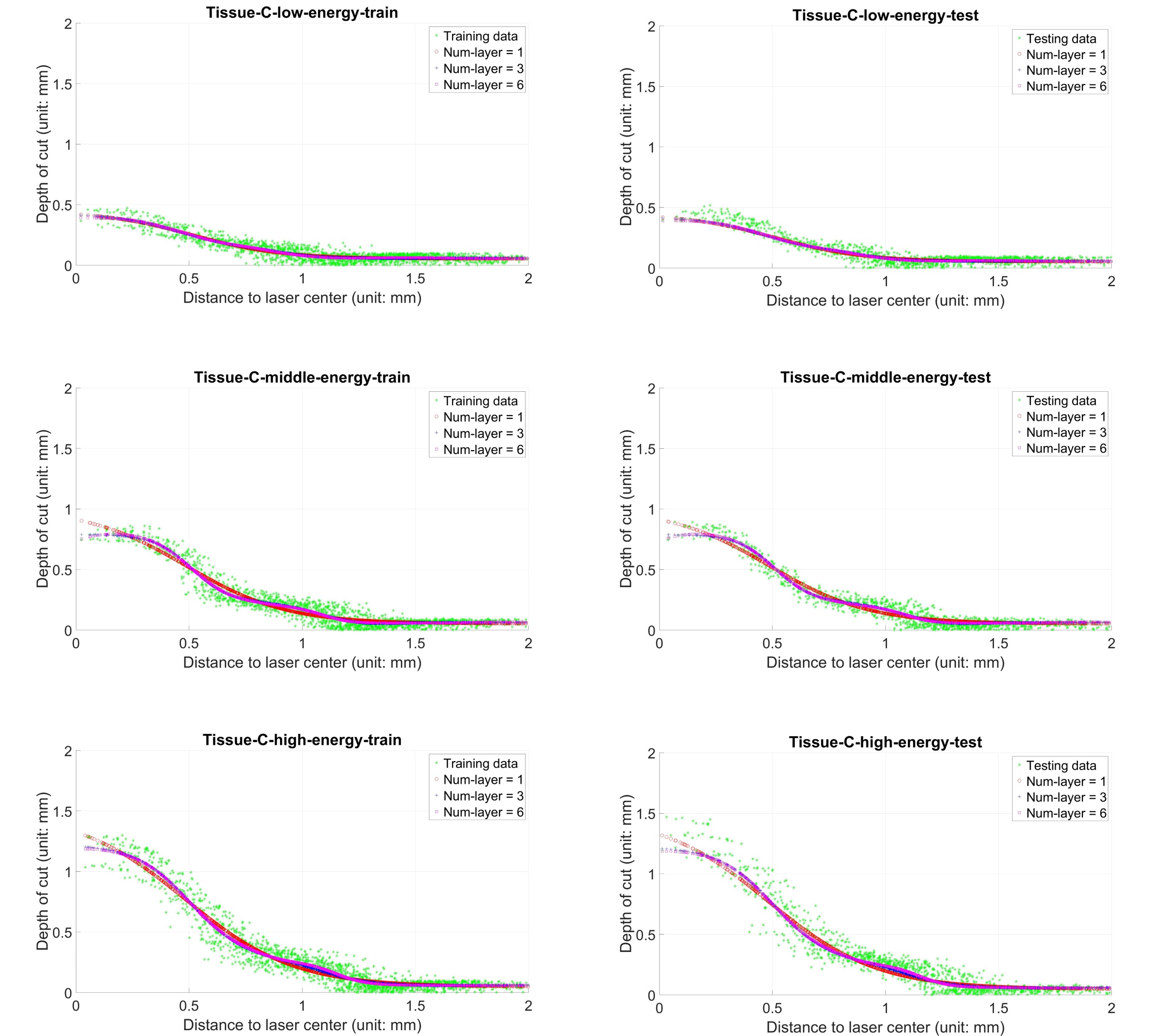}
\caption{Regression analysis of training-testing datasets for Tissue-C ($2.5\%$ agar). The scattered dots depict the data points of the dataset (training as left column and testing as right column). The data labelled with different colors represent the outputs from the neural networks with various layers for regression. }
\label{fig_regression_C}
\end{figure*}

The geometric relation between the distance-to-laser-center and the depth-of-cut is described as a Single layer perceptron model (SLP). For both the training and testing datasets, the regression analysis of Tissue-A (agar-$0.5\%$), Tissue-B (agar-$1.5\%$) and Tissue-C (agar-$2.5\%$) are reported in Fig.~\ref{fig_regression_A}, Fig.~\ref{fig_regression_B} and Fig.~\ref{fig_regression_C} respectively with the low, middle and high energy levels. 

\subsection{Cavity Visualization}

\begin{figure*}[h]
\centering
\includegraphics[scale = 0.50]{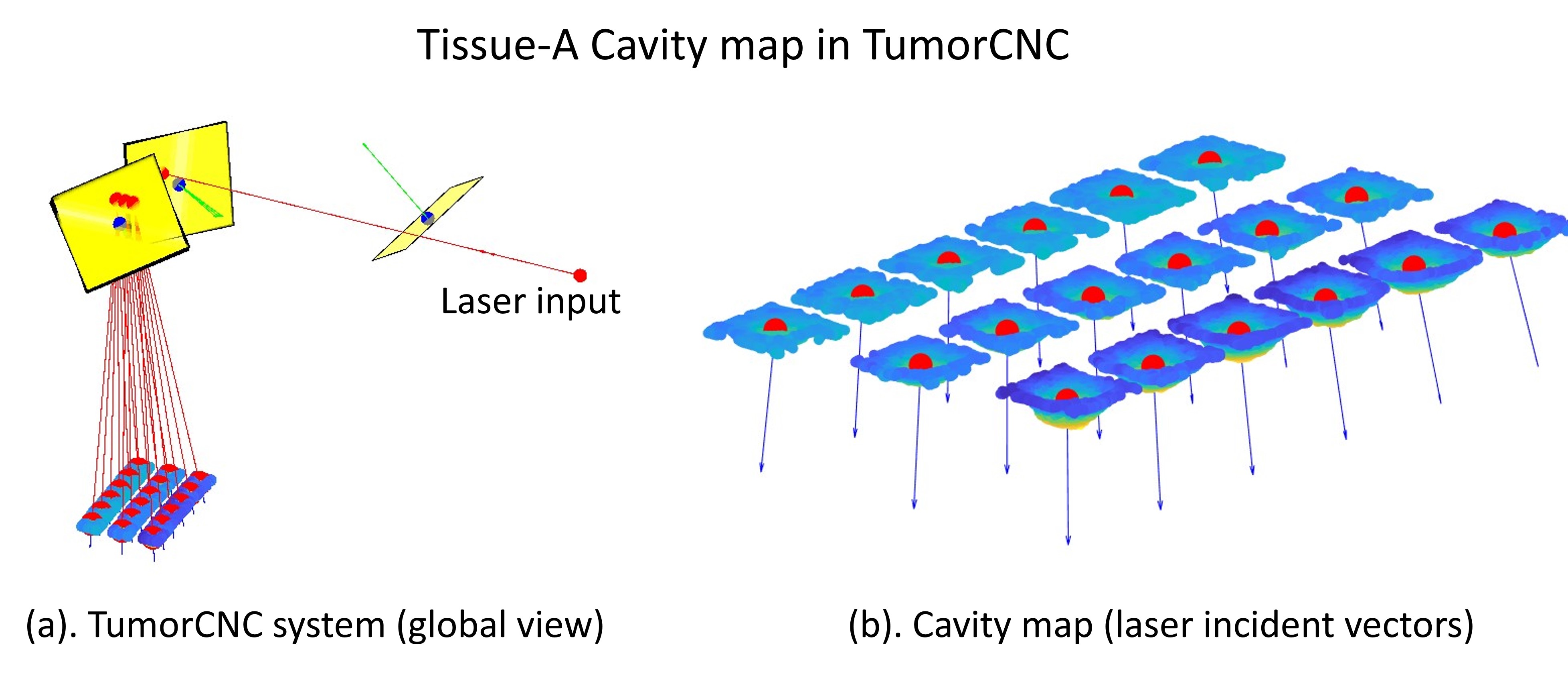}
\caption{Visualization of laser-mirror geometry and ablation cavities (Tissue-A: agar-$0.5\%$).}
\label{cnc_cavity_1}
\end{figure*}

\begin{figure*}[h]
\centering
\includegraphics[scale = 0.50]{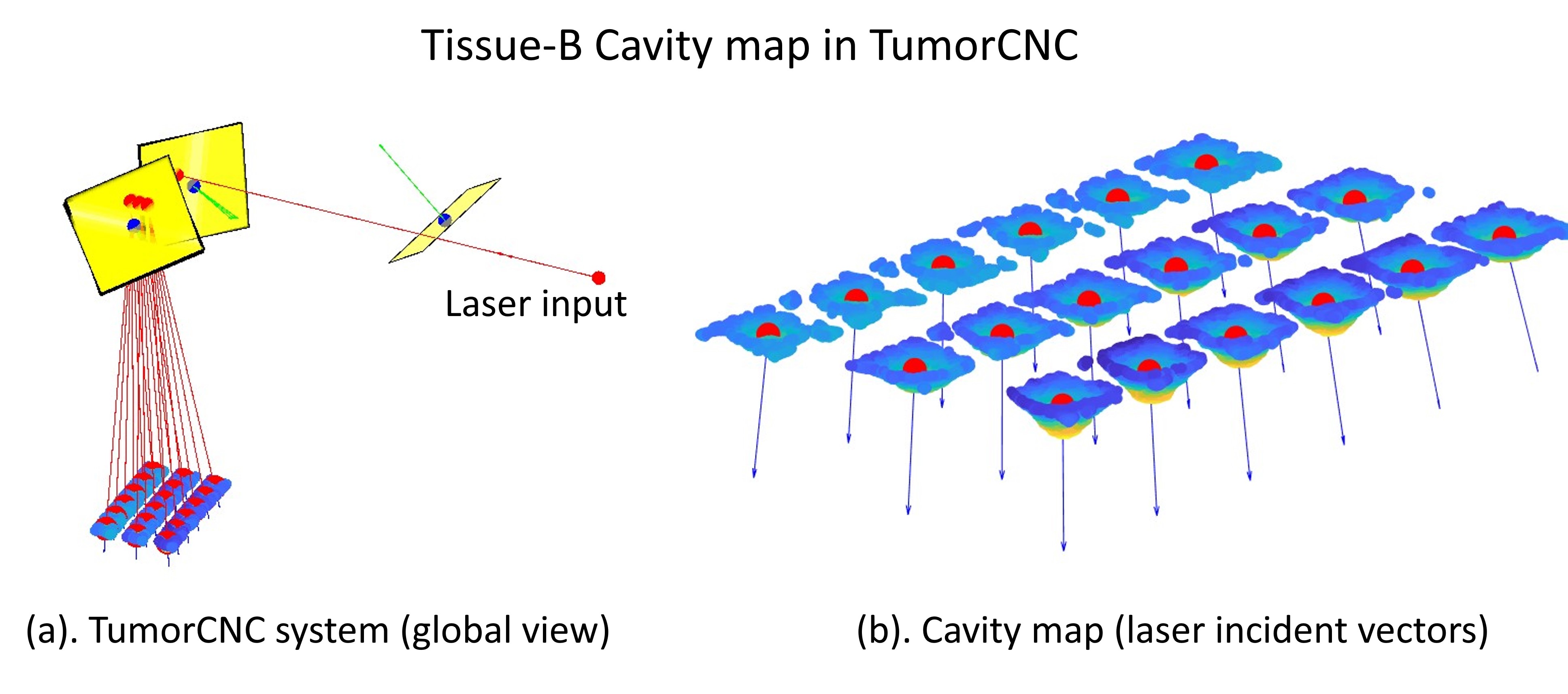}
\caption{Visualization of laser-mirror geometry and ablation cavities (Tissue-B: agar-$1.5\%$).}
\label{cnc_cavity_2}
\end{figure*}

\begin{figure*}[h]
\centering
\includegraphics[scale = 0.50]{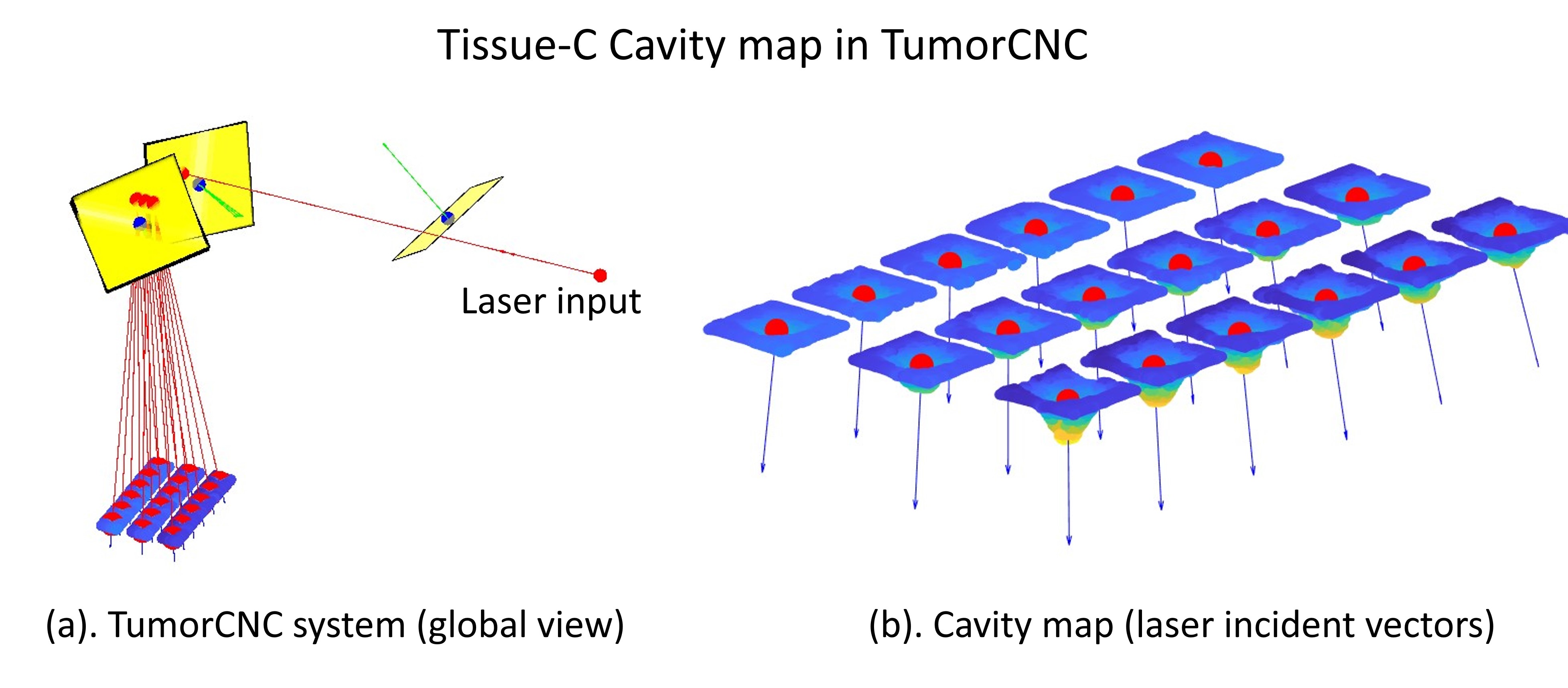}
\caption{Visualization of laser-mirror geometry and ablation cavities (Tissue-C: agar-$2.5\%$).}
\label{cnc_cavity_3}
\end{figure*}

The geometric visualizations of the laser-mirror intersection for the TumorCNC robotic laser system are depicted in Fig.~\ref{cnc_cavity_1} (agar-$0.5\%$), Fig.~\ref{cnc_cavity_2} (agar-$1.5\%$) and Fig.~\ref{cnc_cavity_3} (agar-$2.5\%$). These figures show that the shapes of the cavities are unique based on the properties of the phantom (different ratios of agar for the phantom designs). 

\subsection{Numerical Simulation for the Robotic Planning Problem}

\begin{figure*}[h]
\centering
\includegraphics[scale = 0.80]{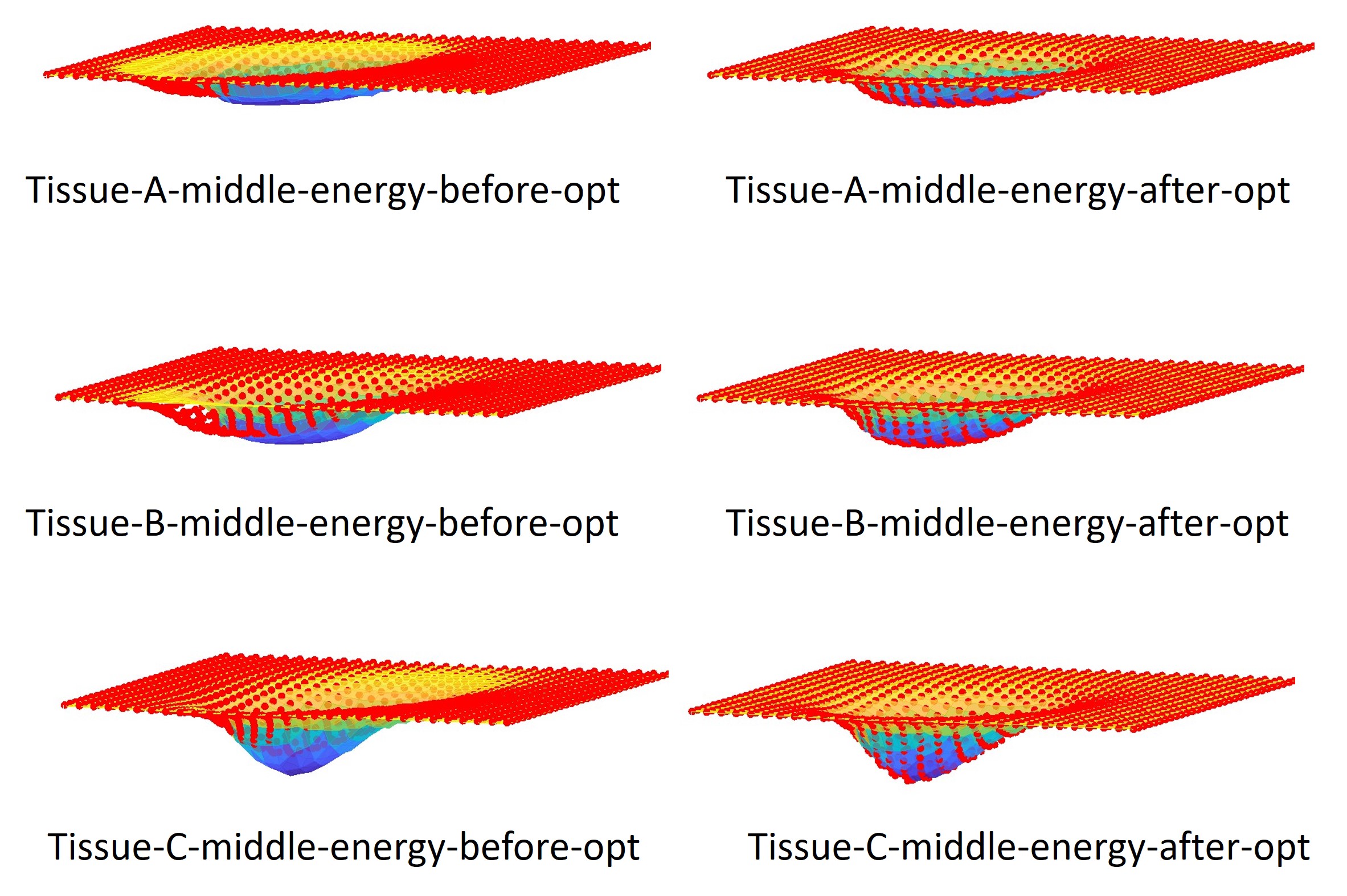}
\caption{Visualization of the predicted and target cavities before and after the optimization (middle energy). The point cloud labelled as red color is the predicted cavity. After optimization, the predicted cavity can be aligned with the target cavity.}
\label{sim_opt_1}
\end{figure*}

\begin{figure*}[h]
\centering
\includegraphics[scale = 0.80]{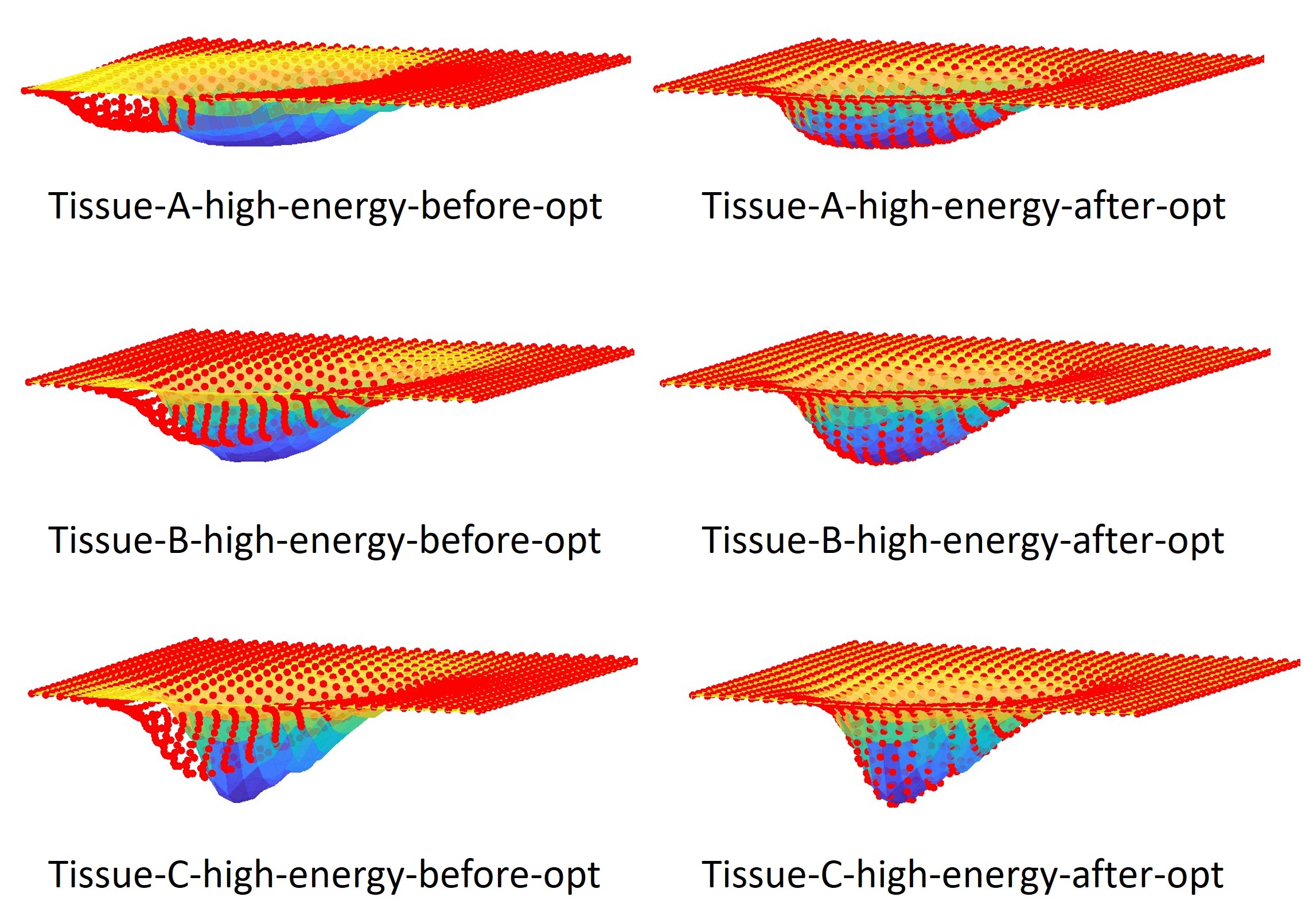}
\caption{Visualization of the predicted and target cavities before and after the optimization (high energy). The point cloud labelled as red color is the predicted cavity. After optimization, the predicted cavity can be aligned with the target cavity.}
\label{sim_opt_2}
\end{figure*}

Fig.~\ref{sim_opt_1} (middle-energy model) and Fig.~\ref{sim_opt_2} (high-energy model) illustrate the shapes of the cavities before and after the optimization. These figures indicate that the offsets between the predicted cavity and the target cavity (ground truth) can be minimized after the optimization. 

\end{document}